\documentclass[a4paper,11pt]{article}
\usepackage[utf8]{inputenc}
\usepackage{placeins}
\usepackage{booktabs}
\usepackage{jheppubMod}
\usepackage{amsmath,amsfonts,amssymb,graphicx,MnSymbol,subfigure}
\usepackage{siunitx}
\usepackage{hyperref} 
\usepackage{tikz}
\usetikzlibrary{shapes.geometric, arrows}
\usepackage[toc,page]{appendix}
\usepackage{adjustbox}
\usepackage{slashed}
\usepackage{multirow}
\usepackage{braket}
\usepackage{physics}
\usepackage{esint}
\usepackage{gensymb}
\usepackage{afterpage}
\usepackage{listings}
\newcommand{\Qtruth}{Q}
\newcommand{\Qpred}{\hat{Q}}
\newcommand{\Jtruth}{J}
\newcommand{\Jpred}{\hat{J}_{\mathrm{RO}}}
\newcommand{\Jgrad}{\hat{J}_{\mathrm{grad}}}
\newcommand{\JdefRO}{J_{\mathrm{RO}}}
\newcommand{\JdefGrad}{J_{\mathrm{grad}}}

\title{A physics-informed uncertainty-aware surrogate for $p$-$n$ junction diodes}

\author[a]{Suchita Kulkarni,}

\affiliation[e]{Institute of Physics, NAWI Graz, University of Graz, Universit\"atsplatz 5, A-8010 Graz, Austria}

\emailAdd{suchita.kulkarni@uni-graz.at}

\abstract{
We develop an uncertainty-aware physics-informed surrogate for a one-dimensional $p$-$n$ junction diode, conditioned on device length, doping and bias. Using the internal state of the device as predicted by the surrogate we develop a test-suite of quantities of interest including the terminal current, depletion width, peak field and stored charge. The formalism and the quantities are can be extended to asymmetric junction and two-dimensional diodes as shown by means of benchmark analysis. For the one-dimensional symmetric junction surrogate, we obtain 0.11\% median $L_1$ error on the electrostatic potential, and 1.5\% and 0.61\% on the electron and hole densities. Finally, we develop an uncertainty analysis which determines the validity of surrogate answer within the user-specified tolerance.}

\begin{document}

\maketitle

\section{Introduction}
Designing a semiconductor device means solving the coupled electrostatic and carrier-transport equations for a given geometry, doping profile and set of boundary conditions. This is routinely done with technology computer-aided design (TCAD) packages such as Synopsys Sentaurus~\cite{Sentaurus_Device_W-2024.09-SP1}, Silvaco Atlas~\cite{silvaco_sftw} and COMSOL Multiphysics~\cite{comsol_sftw}, which solve the underlying equations accurately for one fully specified device and bias. Exploring a design space, however, means many such solves, each started from scratch, over a large space of geometries, doping profiles and operating conditions.

A faster route to explore this space is useful if it reliably returns quantities of interest (QoI) to a device physicist such as the current, the stored charge, the electric field and the depletion width. Physics-informed neural networks~\cite{raissi_physics-informed_2019} are a natural candidate for designing such a surrogate. A physics-informed network represents the internal state of the device, the potential and the carrier densities, and is trained to satisfy the governing equations as well as to match reference data. The QoIs are then derived from this one state, so a new QoI does not require a new model. Because the equations are enforced across the device and not only at labelled points, the network is also constrained between the devices it was trained on, and in principle beyond them. Physics can enter through the architecture, the loss, or the training data. In this work, we test whether and for which quantities, these motivations hold in practice.

Surrogates and physics-informed networks for TCAD have recently attracted considerable attention. Machine-learning approaches to device simulation fall into three groups. Terminal response surrogates map device and bias parameters directly to terminal characteristics such as I--V or C--V curves or figures of merit~\cite{Mehta:2021ae,Myung:2021tcad,Wong:2026aitcad}. They are fast and accurate within their training range but carry no internal state, so every new quantity needs a new model. Physics-informed solvers represent the potential and carrier densities with a network trained on the drift-diffusion residual, with or without reference data~\cite{cao_physics-informed_2023,cai_multi-order_2024,kim_novel_2023,2025arXiv250908073R,Lu:2024pinntcad}. Mesh-native and operator surrogates learn the fields on the TCAD mesh or grid, through graph networks~\cite{popryho2026mesh,2026arXiv260629272Z} or neural operators~\cite{Kim:2024ndd}, and reach realistic two- and three-dimensional geometries. Uncertainty enters these works as ensemble disagreement used to select new simulations~\cite{popryho2026mesh} or as a model-based error bound for a single device~\cite{Zhang:2026sps}. The present work asks what a physics-informed surrogate conditioned over device properties can reliably deliver by utilising conformal methods to estimate uncertainty. 

In this work, we thus intend to answer two related questions, first the capability of a physics-informed approach to deliver QoIs and second, the reliability of those answers at query time when the ground truth is unknown. We construct a physics-informed surrogate conditioned on device parameters, e.g., device length, doping and forward bias, and derive five QoIs -- current, depletion region, electric field, and the stored charges for electrons and holes -- from its predicted internal state. We demonstrate the generalisation of our formalism and architecture by evaluating single-device benchmarks of an asymmetric junction and a two-dimensional planar diode. The full study is carried out for a family of one-dimensional symmetric abrupt $p$-$n$ junction diodes as the minimal complete showcase. We characterise QoIs against the reference simulation and attach an interval that tells a device physicist when the surrogate can be trusted.

We present three findings. First, within our architecture, the terminal current cannot be obtained by differentiating the network's internal state. The derivative must be inversely proportional to the number density across up to fourteen decades within a few Debye lengths wide junction layer. This presents a rather challenging situation for the network. The differentiated current is wrong by about an order of magnitude in long, highly doped devices as the slope fails to follow the required steepness. It even has the wrong sign in two dimensions, although the internal states themselves are accurate. An integral of the potential, the classic Gummel relation~\cite{MollRoss1956,Gummel1970} recovers the current. Second, conditioning the error on the depleted fraction of the device yields intervals for all five QoIs whose coverage holds on held-out devices, whereas the spread of an ensemble of networks covers only about half of the errors. Third, extrapolation in bias degrades beyond the training range and fails beyond the built-in potential, where the device enters a regime absent from the training data.

The paper is organised as follows. Section~\ref{sec:theory} sets up the drift-diffusion system, the current readout and the QoIs. Section~\ref{sec:methods} sets up the methodology we use.  Section~\ref{sec:benchmarks} presents the three single-device benchmarks, which establish that the formalism is not tied to one geometry or doping. Sections~\ref{sec:results} and~\ref{sec:uncertainty} address this question for the symmetric family, through the accuracy of each QoI and its calibrated interval. Section~\ref{sec:conclusion} concludes.

\section{Drift-diffusion system for diodes}
\label{sec:theory}
\subsection{Governing equations and parametrization}
\label{sec:formalism-governing}
We consider a $p$-$n$ junction diode occupying a domain $\Omega\subset\mathbb R^d$ with net doping $N_{\rm net}(\mathbf x)$~\footnote{All boldface quantities indicate d-dimensional element, either a vector or volume depending on the context.}, acceptors $n_A$ on the $p$-side and donors $n_D$ on the $n$-side, and an applied bias $V$. Under the classical drift-diffusion approximation, such a system is characterized by the drift-diffusion equations which involve Poisson's equation and the electron and hole continuity equations, coupled through the carrier transport relations and a recombination model. Here we consider Shockley-Read-Hall (SRH) recombination. The equations are given in Appendix~\ref{app:dd-theory}. We use the well-known quasi-Fermi-level (QFL) parametrization, in which the unknowns are the electrostatic potential and the electron and hole quasi-Fermi potentials, and each carrier current is a single derivative of one of them rather than a balance of drift and diffusion terms that cancel to many orders of magnitude.

We non-dimensionalise the system, with lengths scaled by $\ell$, the longest side of the device ($\ell=L$ in one dimension), and denote the normalized quantities by a tilde ($\tilde\cdot$)~\footnote{Equivalently, we denote all neural network predicted quantities by a `hat' ($\hat\cdot$).}. We follow most definitions from~\cite{2025arXiv250908073R}, and convert them to the QFL formalism where necessary. In Eq.~\eqref{eq:formalism-scaling} we provide the complete list used for non-dimensionalisation.

The non-dimensionalised Poisson's equation and the continuity equations become
\begin{equation}
\lambda^2\,\tilde\nabla^2\tilde\phi = \tilde n-\tilde p-\tilde N_{\mathrm{net}}, \qquad
\tilde\nabla\!\cdot\tilde{\mathbf J}_n = \tilde R, \qquad \tilde\nabla\!\cdot\tilde{\mathbf J}_p = -\tilde R,
\label{eq:formalism-nondim-pdes}
\end{equation}
with the dimensionless potentials $\tilde\phi=\psi/U_t$ and $\tilde v_{n,p}=\phi_{n,p}/U_t$, where $\psi$ is the electrostatic potential and $\phi_{n,p}$ are the quasi-Fermi potentials in volts, $\tilde n = \tilde n_i\exp(\tilde\phi-\tilde v_n)$, $\tilde p=\tilde n_i\exp(\tilde v_p-\tilde\phi)$
and
$\tilde{\mathbf J}_n=-\tilde\mu_n\tilde n\,\tilde\nabla\tilde v_n$, $\tilde{\mathbf J}_p=-\tilde\mu_p\tilde p\,\tilde\nabla\tilde v_p$,
with, 
\begin{equation}
\tilde N_{\mathrm{net}}(\boldsymbol\chi)=\frac{N_{\mathrm{net}}(\mathbf x)}{n_{\max}}\;\longrightarrow\;-\frac{n_A}{n_{\max}}\ \ (p\text{-bulk}),\qquad +\frac{n_D}{n_{\max}}\ \ (n\text{-bulk}),
\label{eq:formalism-Nhat}
\end{equation}
whose larger magnitude is $1$ by construction and whose smaller magnitude is $10^{-|r|}$ with $r=\log_{10}(n_A/n_D)$.  

\subsection{Boundary conditions}
\label{sec:bc}

The boundary $\partial\Omega$ consists of ideal ohmic contacts $\Gamma_c$ and, for $d>1$, of insulating segments. On contact $c$, held at the bias $V_c$ ($V$ on the anode, $0$ on the cathode), charge neutrality and equal quasi-Fermi levels give
\begin{equation}
\tilde v_n=\tilde v_p=\frac{V_c}{U_t},\qquad \tilde\phi=\frac{V_c}{U_t}+\operatorname{asinh}\frac{\tilde N_c}{2\tilde n_i}\qquad\text{on }\Gamma_c,
\label{eq:bc-qfl}
\end{equation}
with $\tilde N_c$ the net doping at the contact. 
On insulating segments
\begin{equation}
\partial_\nu\tilde\phi=0,\qquad \tilde{\mathbf J}_n\!\cdot\!\boldsymbol\nu=\tilde{\mathbf J}_p\!\cdot\!\boldsymbol\nu=0 .
\label{eq:bc-ins}
\end{equation}
The quasi-Fermi boundary values are independent of the doping, which enters the boundary data only through $\tilde\phi$. The built-in potential of a junction is $V_{\rm bi}=U_t\ln(n_A n_D/n_i^2)$, and the potential drop across it at bias $V$ is $V_{\rm bi}-V$. 

\subsection{Terminal current}
\label{sec:readout-current}
The quantity read off a device is the terminal current, the flux of current through the biased contact $\Gamma_a$,
\begin{equation}
J = J_{\rm ref}\,\ell^{d-1}\int_{\Gamma_a}\big(\tilde{\mathbf J}_n+\tilde{\mathbf J}_p\big)\!\cdot\!\boldsymbol\nu\;d\tilde\Gamma ,
\label{eq:QoI_current}
\end{equation}
with $\boldsymbol\nu$ the inward unit normal. $\Gamma_a$ is a point in one dimension and a line segment in two, so $J$ is a current density in 1D and a current per unit depth in 2D. Therefore the current has unit of $\rm{A}/\rm{cm}^2$ in 1D and $\rm{A}/\rm{cm}$ in 2D.  

Setting $U_n=e^{-\tilde v_n}$ and $U_p=e^{\tilde v_p}$ (not to be confused with the thermal voltage $U_t$), the currents become
\begin{equation}
\tilde{\mathbf J}_n=\tilde\mu_n\tilde n_i\,e^{\tilde\phi}\tilde\nabla U_n,\qquad\qquad \tilde{\mathbf J}_p=-\tilde\mu_p\tilde n_i\,e^{-\tilde\phi}\tilde\nabla U_p,
\label{eq:Jdiff}
\end{equation}
and the continuity equations of Eq.~\eqref{eq:formalism-nondim-pdes} become, for both carriers,
\begin{equation}
\tilde\nabla\!\cdot\!\big(a\,\tilde\nabla U\big)=\tilde R ,
\label{eq:readout_current}
\end{equation}
with $U_{n,p}$ fixed on the contacts and $\partial_\nu U=0$ on insulating segments. For given $\tilde\phi$ and $\tilde R$ this is linear in $U$, so no derivative of the network output enters. Here $\tilde\phi$ and $\tilde R$ are the network's, and the end values of $U_{n,p}$ are the network's values on the boundary of the modelled domain, which excludes a thin layer at each contact (Sec.~\ref{sec:dataset}). The readout current is Eq.~\eqref{eq:QoI_current} evaluated with the $\tilde{\mathbf J}$ of this solution.

In one dimension Eq.~\eqref{eq:readout_current} integrates in closed form. The currents are $\tilde J_{n}(\chi)=\tilde J_{n0}+C(\chi)$ and $\tilde J_{p}(\chi)=\tilde J_{p0}-C(\chi)$ with $C(\chi)=\int_{\chi_-}^\chi\tilde R\,d\chi'$, and integrating $dU_n/d\chi=\tilde J_n e^{-\tilde\phi}/(\tilde\mu_n\tilde n_i)$ and $dU_p/d\chi=-\tilde J_p e^{\tilde\phi}/(\tilde\mu_p\tilde n_i)$ across the modelled domain $[\chi_-,\chi_+]$ gives the currents at its first end,
\begin{align}
\tilde J_{n0}&=\frac{U_n(\chi_+)-U_n(\chi_-)-\int_{\chi_-}^{\chi_+} C\,e^{-\tilde\phi}/(\tilde\mu_n\tilde n_i)\,d\chi}{\int_{\chi_-}^{\chi_+} e^{-\tilde\phi}/(\tilde\mu_n\tilde n_i)\,d\chi},\\
\tilde J_{p0}&=-\frac{U_p(\chi_+)-U_p(\chi_-)-\int_{\chi_-}^{\chi_+} C\,e^{\tilde\phi}/(\tilde\mu_p\tilde n_i)\,d\chi}{\int_{\chi_-}^{\chi_+} e^{\tilde\phi}/(\tilde\mu_p\tilde n_i)\,d\chi}.
\label{eq:deriv_free_current}
\end{align}
The total $\tilde J_{n0}+\tilde J_{p0}$ is the terminal current, constant in $\chi$ because the recombination integrals cancel. For $d>1$ we solve Eq.~\eqref{eq:readout_current} by cell-centred finite volumes on a uniform grid of the modelled domain. Insulating faces carry no flux, which imposes $\partial_\nu U=0$ without further input. This is a post-processing step at evaluation time that takes the place of the 1D quadrature. 

Eq.~\eqref{eq:deriv_free_current} involves only $\tilde\phi$, $\tilde R$, $\tilde\mu_{n,p}$, $\tilde n_i$ and the end values of $U_{n,p}$, and the quasi-Fermi boundary values of Eq.~\eqref{eq:bc-qfl} do not depend on $n_A$ or $n_D$. The formulation therefore holds unchanged for $n_A\neq n_D$ and in any dimension. 

\subsection{Quantities of interest and scoring metrics}
\label{sec:qoi}
Given the network output, we derive five scalar QoIs, the terminal current $\Jpred$, the depletion width $W$, the peak field $\mathcal{E}_{\rm peak}$ and the stored charges $Q_n$, $Q_p$. They are the quantities for which Sec.~\ref{sec:uncertainty} constructs an uncertainty. Additionally, we also consider $\psi$, $n$, $p$, $R$ and $\Delta E_F$ as fields over $\Omega$. Finally, as said before we compute $\Jgrad$, but do not actively use it in our analysis. Below we define each of the QoI that we use.

\begin{enumerate}
    \item \textbf{Terminal current.} We define two estimates. The differentiated current
    \begin{equation}
    \JdefGrad = J_{\rm ref}\,\ell^{d-1}\!\int_{\Gamma_a}\!\bigl(-\tilde\mu_n\tilde n\,\tilde\nabla\tilde v_n - \tilde\mu_p\tilde p\,\tilde\nabla\tilde v_p\bigr)\cdot\boldsymbol\nu\;d\tilde\Gamma ,
    \label{eq:Jgrad_def}
    \end{equation}
    with $\tilde n$ and $\tilde p$ as defined above, evaluates the drift-diffusion flux; on the network output the gradients come from automatic differentiation. The readout current
    \begin{equation}
    \JdefRO = J_{\rm ref}\,\ell^{d-1}\!\int_{\Gamma_a}\!\bigl(\tilde\mu_n\tilde n_i e^{\tilde\phi}\tilde\nabla U_n - \tilde\mu_p\tilde n_i e^{-\tilde\phi}\tilde\nabla U_p\bigr)\cdot\boldsymbol\nu\;d\tilde\Gamma ,
    \label{eq:Jpred_def}
    \end{equation}
    evaluates the same integral with the Slotboom fluxes from Eq.~\eqref{eq:readout_current}, which in one dimension reduces to the closed form of Eqs.~\eqref{eq:readout_current} and~\eqref{eq:deriv_free_current}. Throughout this work, we use $\Jpred$ as the terminal-current predicted by the network as QoI. We also evaluate $\Jgrad$ and compare it with $\Jpred$ in App.~\ref{app:derivative}. 
    \item Peak electric field helps quantify the maximum local electrostatic stress and assess field redistribution with applied bias. It is defined as 
    \begin{equation}
\boldsymbol{\mathcal{E}}(\mathbf x) = -\frac{U_t}{\ell}\tilde\nabla\tilde\phi, \qquad
\mathcal{E}_{\mathrm{peak}} = \max_{\mathbf x\in\Omega}|\boldsymbol{\mathcal{E}}(\mathbf x)|.
\label{eq:QoI-efield}
\end{equation}

\item The depletion width sets the junction capacitance, $C_j\approx\varepsilon/W$ per unit area in the depletion approximation. A real doping profile has no sharp edge, therefore we define the depletion region as the set $D$ where $|\boldsymbol{\mathcal{E}}|$ exceeds 1\% of its own peak, and its size $W=|D|$ (the width in one dimension, the area in two),
\begin{equation}
D=\big\{\mathbf x\in\Omega:\ |\boldsymbol{\mathcal{E}}(\mathbf x)|\ge0.01\,\mathcal{E}_{\mathrm{peak}}\big\},\qquad W=|D| .
\label{eq:QoI-depletion}
\end{equation}
The threshold is a modelling choice. For orientation, the abrupt-junction depletion approximation (Eq.~\eqref{eq:depl_width}) places most of the depletion region on the lighter-doped side. 
\item 
The stored charges $Q_n, Q_p$ quantify electron, hole injection and charge storage, which influence transient response and switching behaviour.
The stored charges are the carriers in excess of their equilibrium minority densities, summed over the quasi-neutral part of each side, that is, the part of the doped region that is not depleted. We denote these regions by
\begin{equation}
\Omega_p^{\rm qn}=\{\mathbf x\in\Omega:\ N_{\rm net}(\mathbf x)<0,\ \mathbf x\notin D\},\qquad
\Omega_n^{\rm qn}=\{\mathbf x\in\Omega:\ N_{\rm net}(\mathbf x)>0,\ \mathbf x\notin D\},
\end{equation}
with $D$ from Eq.~\eqref{eq:QoI-depletion}, and write
\begin{align}
Q_{n,\mathrm{stored}} &= q\int_{\Omega_p^{\rm qn}} \max\big(n - n_{p0},\ 0\big)\, d\Omega, \\
Q_{p,\mathrm{stored}} &= q\int_{\Omega_n^{\rm qn}} \max\big(p - p_{n0},\ 0\big)\, d\Omega.
\label{eq:QoI-qstored}
\end{align}
Here $n_{p0}=n_i^2/n_A$ and $p_{n0}=n_i^2/n_D$ are the equilibrium minority densities, with $n_A$ and $n_D$ the acceptor and donor doping of the respective plateau ($n_{p0}=p_{n0}$ for a symmetric device). The measure $d\Omega$ is a length in one dimension and an area in two, so $Q$ is a charge per unit area in 1D and per unit depth in 2D.
The depletion region of Eq.~\eqref{eq:QoI-depletion} is computed for ground truth and network predictions independently.
\item Finally, we define the quasi-Fermi-level splitting  $\Delta E_F(\mathbf x)$. It is the local separation of the two quasi-Fermi potentials,
\begin{equation}
\Delta E_F(\mathbf x) = U_t\ln\!\left(\frac{n(\mathbf x)\,p(\mathbf x)}{n_i^2}\right) = U_t\big(\tilde v_p(\boldsymbol\chi) - \tilde v_n(\boldsymbol\chi)\big).
\label{eq:QoI-def-qfl}
\end{equation}
\end{enumerate}

We therefore have five scalar quantities, and we need to establish the accuracy of network prediction. A scalar QoI $Q$ is scored at each (device, voltage) point by the relative error of the surrogate answer $\Qpred$ against ground truth,
\begin{equation}
e = \left|\frac{\Qpred}{\Qtruth}-1\right|,
\label{eq:QoI-e}
\end{equation}
and a profile by its relative $L_1$ error $\|\hat q-q\|_1/\|q\|_1$ over $\Omega$. Over the surrogate dimensions we consider, not all quantities may always be measured, e.g., depletion width is not defined for a fully depleted device, or an extremely tiny current offers no practical use. We therefore compute the relative error only where the corresponding QoI or profile is defined. The definition uses only the network's own output, and the current is measurable if the predicted current satisfies $|\Jpred|\ge10^{-6}$\,A/cm$^2$, the depletion width is informative if the network's $W$ is below 95\% of the domain, and a stored charge is informative if the network's charge is positive. Finally, each quantity is evaluated on the domain it is defined on, which means the ground truth is evaluated over the entire device dimensions and for reasons explained below the network predictions are evaluated over a slightly smaller range. 

\section{Methods}
\label{sec:methods}
\subsection{Dataset}
\label{sec:dataset}

Ground truth is generated with the open-source TCAD solver DEVSIM~\footnote{We use DEVSIM as an open source alternative to the industrial TCAD solvers. For uncertainty analysis we generate multiple custom datasets and thus using DEVSIM is preferred over open source datasets.}~\cite{Sanchez2022}, solving the steady-state drift--diffusion system for a silicon p--n diode at $T=300$\,K ($U_t=25.89$\,mV, $n_i=10^{10}$\,cm$^{-3}$, $\varepsilon=11.7\,\varepsilon_0$) with constant mobilities $\mu_n=1417$ and $\mu_p=470.5$\,cm$^2$/V\,s and equal lifetimes $\tau_n=\tau_p=10^{-5}$\,s~\cite{2025arXiv250908073R}. The doping is smeared by a Gaussian kernel with standard deviation $w/\sqrt2$, with $w=0.02\,L$,
\begin{equation}
N_{\mathrm{net}}(x)=\frac{n_D-n_A}{2}+\frac{n_D+n_A}{2}\,\mathrm{erf}\!\left(\frac{x-x_j}{w}\right),
\label{eq:doping-general}
\end{equation}
which tends to $-n_A$ on the $p$-side and $+n_D$ on the $n$-side. 

The profiles are interpolated by cubic splines in $(\tilde\phi,\ln n,\ln p)$ in one dimension and by a $C^1$ Clough–Tocher interpolant on the mesh nodes in two and converted to $(\tilde v_n,\tilde v_p)$ via the Boltzmann relations of Sec.~\ref{sec:formalism-governing}. We observe that DEVSIM forces exact equilibrium on the contact node itself, which does not join smoothly onto the injected minority-carrier profile one cell inside, leading to a particularly fast-varying profile near contacts. 

This fast varying profile is a particularly challenging feature of the dataset and the underlying physics. As it was preempted in sec.~\ref{sec:readout-current}, we discuss our approach to address this challenge. At each contact the minority density rises sharply from its equilibrium value, so that the quasi-Fermi level of the minority carrier leaves its contact value within a few nanometres. This sharp change within a few nanometers of contact is very challenging to be reproduced by the network.  We explicitly quantify the error introduced due to this trim within our analysis. 

We therefore trim 1 nm of data for 1D devices and 1.4 nm of data for 2D devices. The trim makes the device 2\,nm, 2.8\, nm shorter in 1D, 2D, so a quantity computed on the trimmed window differs from the same quantity computed on the full device even for a perfect network. We measure the error introduced by this approximation for the 1D symmetric devices by computing the QoIs from DEVSIM's own solution twice, once on the full length $L$ and once on $[1\,\mathrm{nm},L-1\,\mathrm{nm}]$. The QoIs lose at most 1\% at the 95th percentile. The largest losses are 1.5\% for $W$, 1.4\% for $Q_p$, 0.25\% for the readout current, 0.0045\% for the peak field and 4.2\% for $Q_n$. They are concentrated in the devices shorter than $0.4\,\mu$m, where the 2\,nm trim is up to 2\% of the length. In the longer devices the largest losses are less than percent.

Our predicted quantities $\hat Q$ are therefore always computed on the trimmed domain, while the ground truth is always computed on the full device. The errors therefore also include any effect of the trim.

\subsubsection{Benchmark data sets}
\label{sec:bench-data}
\begin{figure}[htbp]
	\centering
	\includegraphics[width=0.5\textwidth]{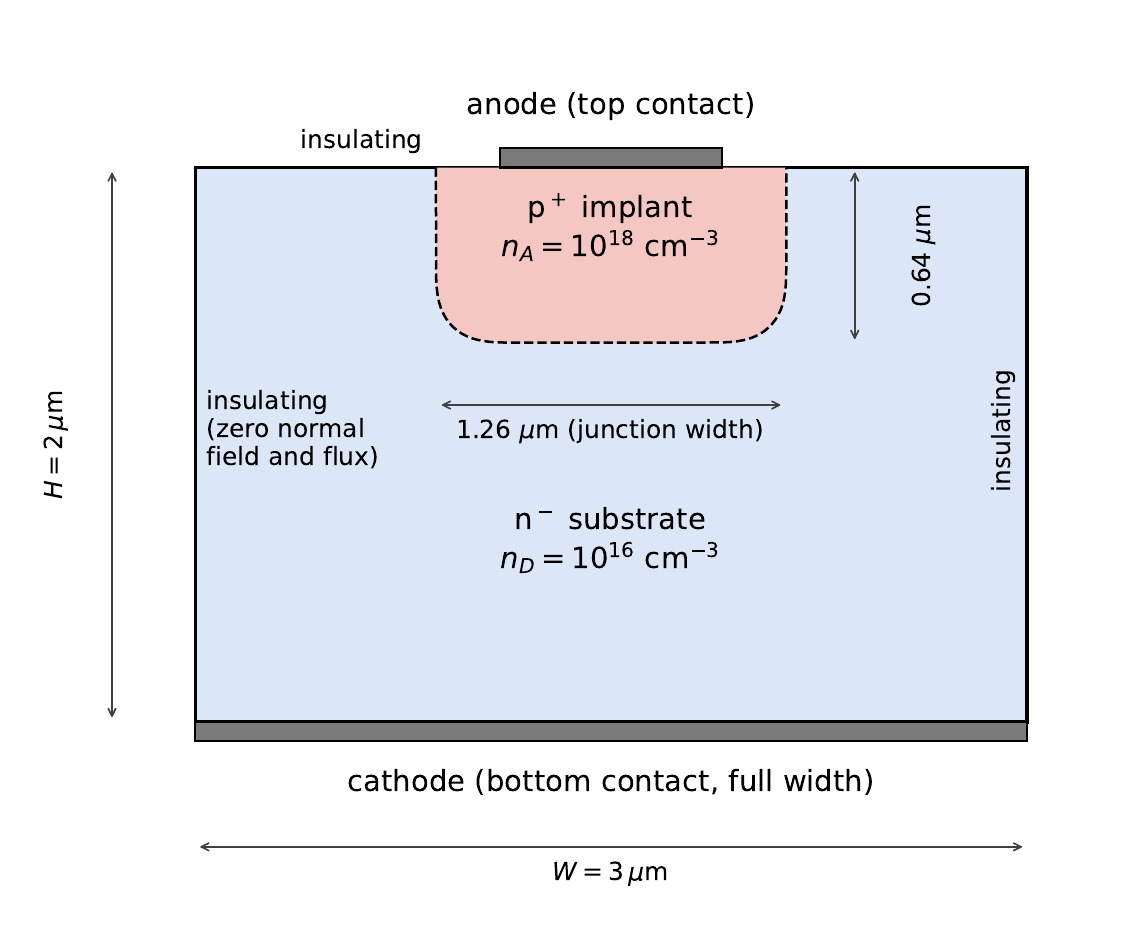}
	\caption{The 2D planar benchmark device (contact thickness exaggerated): a p$^+$ implant ($n_A=10^{18}$\,cm$^{-3}$, 1.26\,$\mu$m wide, 0.64\,$\mu$m deep) in an n$^-$ substrate ($n_D=10^{16}$\,cm$^{-3}$) of width 3\,$\mu$m and height 2\,$\mu$m. The anode is a window on the top face, the cathode covers the bottom face, and the remaining boundary is insulating.}
	\label{fig:planar2d-geometry}
\end{figure}
Besides the one-dimensional symmetric data set of Sec.~\ref{sec:dataset}, on which the detailed study of Secs.~\ref{sec:results} and~\ref{sec:uncertainty} rests, we show the framework applicability on three single-device benchmarks, where each device is solved for 16 forward voltages. Twelve are used for training jointly across all voltages and four are held-out for evaluation. Both 1D devices have $L = 1.80\,\mu\rm{m}$: the symmetric device has $n_A=n_D = 2.91\times10^{17}\, \rm{cm}^{-3}$, and the asymmetric device has $n_A=6\times10^{17}\,\rm{cm}^{-3}$ and $n_D=6\times10^{16}\,\rm{cm}^{-3}$ ($n_A/n_D=10$). For  2D device we set $n_D=10^{16}\, \rm{cm}^{-3}$ with the detailed geometry given in fig.~\ref{fig:planar2d-geometry}. 

For the 2D benchmark a quarter of the collocation and data points are drawn near the metallurgical junction. As in 1D, the modelled domain excludes the first two mesh rows for 2D devices. The benchmarks establish that the formalism and the readout are generic. A single-device fit, however, says nothing about devices the network has not seen. It requires a device family, held-out devices and calibration data, which we construct for the symmetric 1D diode in Secs.~\ref{sec:results}--\ref{sec:uncertainty} .
\subsubsection{1D symmetric surrogate dataset}
\label{sec:1D-data}
For a 1D surrogate construction, we consider a family of devices which vary in length and doping concentration. Devices are drawn over the range of $L\in[10^{-5},7\times10^{-4}]$\,cm, $n_D\in[10^{15},10^{19}]$\,cm$^{-3}$~\footnote{The classical drift-diffusion model has limitations at high doping, where constant mobility and Boltzmann statistics may become inadequate. Our focus here is on reproducing DEVSIM reference solutions. Inforporating more realistic high-doping physics is left for future work.}. The upper bound on length is the largest length at which DEVSIM converged reliably under bias. Each device is solved at the forward voltages 0.04 to $V/V_{bi}<0.8$ (with $V_{bi}$ as in Sec.~\ref{sec:bc}) in steps of $0.04$ V and up to a maximum of 0.64\,V, so devices with $n_D\gtrsim 5\times10^{16}$\,cm$^{-3}$ reach only $0.6$--$0.8\,V_{bi}$. We dub one device--voltage pair as one anchor.  We generate three datasets with each serving a specific role defined in Table~\ref{tab:datasets}.

\begin{table}[htbp]
\centering

\setlength{\tabcolsep}{3pt}
\begin{tabular}{@{}|l|l|c|@{}}
\hline
Set & Role & Devices (anchors) \\
\hline
Training & fit weights & 27 (404) \\
Held-out & validation & 3 (45)  \\
Calibration & analysis, intervals & 310 (4586)\\
Test & all reported claims & 246 (3647) \\
Extrapolation & $V$ extrapolation & 12 (---) \\
\hline
\end{tabular}
\caption{Device sets and their roles. Anchors are counted after restricting to $V/V_{bi}<0.8$ and, for the calibration and test sets, to the supported range $\kappa\ge-2.5$ of Sec.~\ref{sec:uncertainty-framework}.}
\label{tab:datasets}
\end{table}

The first dataset consists of the 30 devices or 449 anchors with $V/V_{bi}<0.8$, drawn using Latin hypercube sampling over $\log_{10}(n_D),\log_{10}(L)$. Three whole devices are withheld at the training time. A separate set of 310 devices is used to develop the analysis and to calibrate the error intervals of Sec.~\ref{sec:uncertainty}. The test set consists of 250 devices generated with Latin hypercube sampling in $\log_{10}(n_D),\log_{10}(L)$ to evaluate final network performance. It was drawn, solved and evaluated after the networks, the calibration set, the error bars and the analysis were frozen, and it was scored once. Two of the 250 devices did not converge at the first voltage and are not part of the set. The remaining 248 devices are restricted as described above, which leaves 246. Finally, to test extrapolation, we generated additional 12 devices beyond the training range.

With $L, n_D$ spanning about (1.8, 4) decades, the sampled devices cover $L/\lambda_D$ from 1.5 to 3500 (the corners of the box $(L,n_D)$ would give 0.8 to 5400). This includes devices only a Debye length or two long, in which the depletion region occupies the whole device, to long devices with a quasi-neutral bulk. Whether a device counts as fully depleted depends on how the depletion edge is defined, so we report results by the continuous depletion fraction $W_{\mathrm{dep}}/L$ rather than by a binary label.

As we set $w=0.02L$, the doping transition is always resolved on the same fraction of the length, but junction smearing width does not scale with the depletion width. At $n_D=10^{19}$\,cm$^{-3}$ the zero-bias abrupt-junction depletion width is $\approx17$\,nm, whereas $w=140$\,nm at the upper limit of the length range. This shows the variety of devices we consider in our surrogate.
\subsection{Network architecture}
\label{sec:architecture}

\begin{figure}[htbp]
  \centering
  \includegraphics[width=0.9\textwidth]{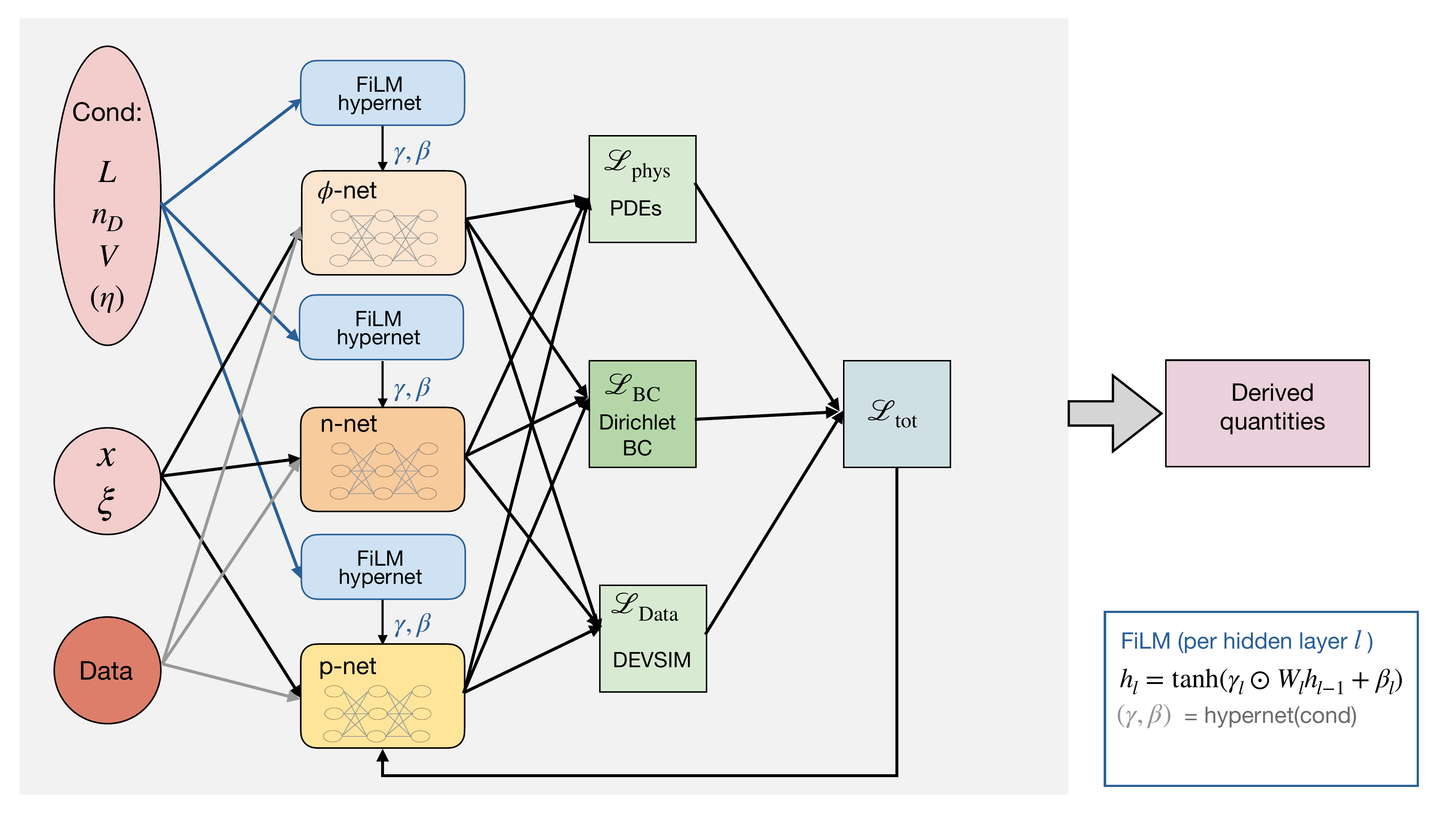}
  \caption{Network architecture representation}
  \label{fig:architecture}
\end{figure}

Our architecture and training methodology is inspired by~\cite{2025arXiv250908073R}. Three separate networks, $\phi$-net, $n$-net and $p$-net, produce $\tilde\phi$, $\tilde v_n$ and $\tilde v_p$ at a point $\boldsymbol\chi=(\mathbf x-\mathbf x_c)/\ell$ of a device of dimension $d$, where $\mathbf x_c$ is the centre of the device and $\ell$ its longest side, so that $\ell=L$ and $\chi=x/L-1/2\in[-1/2, 1/2]$ in one dimension. The full network architecture including the FiLM conditioning described below is shown in fig.~\ref{fig:architecture}. Each is a $\tanh$ multilayer perceptron with five hidden layers of 128 units and a linear read-out. The spatial input has $2d$ components, $\boldsymbol\chi$ and $\boldsymbol\xi=\boldsymbol\chi/\lambda$ with $\lambda=\lambda_D/\ell$. We found $\xi$ to be a very useful variable because it lets the network see the position in units of the Debye length, which changes by orders of magnitude across the grid. 

The device and bias enter through feature-wise linear modulation of every hidden layer. For each of the three networks, a small hypernetwork with one hidden layer of 32 $\tanh$ units maps the conditioning vector $c$ to a scale $\gamma_l=1+\Delta\gamma_l$ and a shift $\beta_l$ for each of the five hidden layers (width 128), and the layer computes
\begin{equation}
h_l=\tanh\!\big(\gamma_l\odot(W_lh_{l-1}+b_l)+\beta_l\big),\qquad h_0=(\boldsymbol\chi,\boldsymbol\xi).
\end{equation}
The last hypernetwork layer is initialised to zero, so training starts from three unconditioned networks ($\gamma=1,\beta=0$). The conditioning vector is
\begin{equation}
c=\Big(\log_{10}(\sqrt{n_An_D}/n^{\rm ref}_D),\ \ \log_{10}(\ell/L^{\rm ref}),\ \ \log_{10}(\mu_p/\mu_n),\ \ V/U_t-V_c,\ \ s,\ \ r,\ \ \mathbf g\Big),  \nonumber
\end{equation}
with $r=\log_{10}(n_A/n_D)$, and for $d=2$, five geometry descriptors $\mathbf g$. These are the log aspect ratio $\log_{10}(L_x/L_y)$ of the domain, the fraction of its face covered by the anode and by the cathode contact, and the two components of the offset of the metallurgical-junction centroid from the domain centre in units of $\ell$. In one dimension $\mathbf g$ is absent.

Among the scalar components, $n^{\rm{ref}}_D = 10^{15}\,\mathrm{cm^{-3}}$, $L^{\rm{ref}} = 3.97\times10^{-4}\,\mathrm{cm}$, $s=\pm1$ marks which side of the device is donor-doped, and $r$ is the doping asymmetry. The mobility ratio and $s$ are constant in this work, and $r=0$ in the symmetric data of Secs.~\ref{sec:results-state}--\ref{sec:results-extrap}, whose networks were trained without this column and therefore have five hypernetwork inputs. The column $V/U_t-V_c$ subtracts a fixed $V_c$, which represents a mean over applied bias $V/U_t$. 

The final simplification concerns the offset of the hole quasi-Fermi level. The quasi-Fermi level $\tilde v_n, \tilde v_p$ profiles behave such that $\tilde v_n = 0, \tilde v_p = V/U_t$ for most of the device, while they change to $\tilde v_n = V/U_t, \, \tilde v_p = 0$ only at the contacts, where they change rapidly. The network therefore must predict a large, nearly constant offset at every point for $\tilde v_p$, on top of a spatial variation of only a few units. Secondly, our 1nm trim removes the layer in which $\tilde v_p$ falls to zero. Therefore, instead of forcing the network to reconstruct this level, the $p$-network returns $f_p(\boldsymbol\chi,\boldsymbol\xi;c)$ and the model sets $\tilde v_p(\boldsymbol\chi) = f_p(\boldsymbol\chi,\boldsymbol\xi;c)+V/U_t$. After the shift, $f_p$ and $\tilde v_n$ have the same range and nearly the same shape up to a sign, each with a flat plateau at zero. 
\subsection{Loss functions}
\label{sec:loss}
We reconstruct carrier densities from the outputs as
\begin{equation}
\tilde n=\tilde n_i\exp\!\big(\mathrm{clip}(\tilde\phi-\tilde v_n,\pm40)\big),\qquad
\tilde p=\tilde n_i\exp\!\big(\mathrm{clip}(\tilde v_p-\tilde\phi,\pm40)\big),
\end{equation}
(the clip at $\pm40$ is a float32 overflow guard). Using definitions in Eq.~\eqref{eq:formalism-nondim-pdes} and Eq.~\eqref{eq:formalism-srh}, the pointwise residuals are
\begin{align}
r_P &= \lambda^2\tilde\nabla^2\tilde\phi-\big(\tilde n-\tilde p-\tilde N_{\mathrm{net}}\big),\label{eq:residuals}\\
r_n &= \frac{1}{{\tilde\mu_n}}\left(\tilde\nabla\!\cdot\tilde{\mathbf J}_n-\tilde R\right),\qquad
r_p = \frac{1}{{\tilde\mu_p}}\left(\tilde\nabla\!\cdot\tilde{\mathbf J}_p+\tilde R\right),\nonumber
\end{align}
where $\tilde N_{\mathrm{net}}$ is the analytic smeared doping above. 

For a minibatch $\mathcal B$ of anchors, with $N=256$ collocation points $\boldsymbol\chi_{a,i}$ per anchor, the terms are mean-squared values over all points of the minibatch
\begin{align}
&\mathcal L_{\mathrm{Poisson}} = \big\langle r_P^2\big\rangle,\\ 
&\mathcal L_{\mathrm{cont}} = \big\langle r_n^2\big\rangle+\big\langle r_p^2\big\rangle,\\
&\mathcal L_{\mathrm{data}} = \big\langle(\tilde\phi-\tilde\phi^{D})^2\big\rangle+\big\langle(\tilde v_n-\tilde v_n^{D})^2\big\rangle+\big\langle(\tilde v_p-\tilde v_p^{D})^2\big\rangle,\\
&\mathcal L_{\mathrm{D}} = \big\langle(\tilde\phi-\tilde\phi^{D})^2\big\rangle_{D}+\big\langle(\tilde v_n-\tilde v_n^{D})^2\big\rangle_{D} + \big\langle(\tilde v_p-\tilde v_p^{D})^2\big\rangle_{D}, \\
&\mathcal L_{\rm N} = \langle(\partial_\nu\tilde\phi)^2\rangle + \langle(\tilde{\mathbf J}_n\!\cdot\!\boldsymbol\nu)^2\rangle+\langle(\tilde{\mathbf J}_p\!\cdot\!\boldsymbol\nu)^2\rangle,
\end{align}
where $^{D}$ denotes the DEVSIM value, $\langle\cdot\rangle$ averages over the collocation points and $\langle\cdot\rangle_{D}$ over the contact points of each anchor. The total loss is
\begin{equation}
    \mathcal{L} = \lambda_{\rm{data}} \mathcal{L}_{\rm{data}} + \lambda_{\rm{D}} \mathcal{L}_{\rm{D}} + \lambda_{\rm{N}} \mathcal{L}_{\rm{N}}  + \lambda_{\rm{Poisson}} \mathcal{L}_{\rm{Poisson}}  + \lambda_{\rm{cont}} \mathcal{L}_{\rm{cont}},
\end{equation}

with all weights equal to 1 without any delay, ramp or curriculum learning. The data term is evaluated at the same points as the collocation points drawn uniformly on the trimmed domain. For $d>1$ a term $\mathcal L_{\rm N}$ imposes Eq.~\eqref{eq:bc-ins} on 128 points per anchor on the insulating segments of the boundary of the modelled domain, with weight 1. Both kinds of boundary condition are therefore soft. The Dirichlet values on the contacts enter through $\mathcal L_{\rm D}$ with DEVSIM's values at the contact points of the trimmed boundary. The Neumann conditions enter through $\mathcal L_{\rm N}$. 

\subsubsection{Density gating of the physics loss}
\label{sec:formalism-gating}
Our training procedure involves minimizing loss via gradient descent, for which the gradient has to flow back through Boltzmann relations. This introduces density gating in loss minimization, in other words below some value of the electron and hole densities the loss becomes insensitive to their change and hence the network can no longer converge to the right solution. We discuss this in some details below, which justifies our choice of augmenting the network with DEVSIM generated data. 

The residual $r_n$ contains $\tilde n$ through $\tilde{\mathbf J}_n$, even though neither $\tilde n$ nor $\tilde p$ appears explicitly in the loss formula $\langle r_n^2\rangle$.

Gradient descent requires $\partial\mathcal{L}_{\mathrm{cont}}/\partial\tilde v_n$. Since $\mathcal{L}_{\mathrm{cont}}=\langle r_n^2\rangle+\langle r_p^2\rangle$,
\begin{equation}
\frac{\partial\mathcal{L}_{\mathrm{cont}}}{\partial\tilde v_n}
= \frac{\partial\langle r_n^2\rangle}{\partial\tilde v_n}+\frac{\partial\langle r_p^2\rangle}{\partial\tilde v_n}.
\end{equation}
The average $\langle\cdot\rangle$ is taken over fixed collocation points, therefore,
\begin{equation}
\frac{\partial\langle r_n^2\rangle}{\partial\tilde v_n}
= \left\langle\frac{\partial r_n^2}{\partial\tilde v_n}\right\rangle
= 2\!\left\langle r_n\,\frac{\partial r_n}{\partial\tilde v_n}\right\rangle.
\end{equation}
Differentiating $r_n$ with respect to $\tilde v_n$, and keeping the dominant terms gives,
\begin{equation}
\frac{\partial r_n}{\partial\tilde v_n}
= \frac{1}{\tilde\mu_n}\,\tilde\nabla\!\cdot\!\left(\frac{\partial\tilde{\mathbf J}_n}{\partial\tilde v_n}\right).
\end{equation}
It therefore suffices to compute $\partial\tilde{\mathbf J}_n/\partial\tilde v_n$ which gives
\begin{align}
\frac{\partial\tilde{\mathbf J}_n}{\partial\tilde v_n}
&= -\tilde\mu_n\,\tilde n\,\frac{\partial}{\partial\tilde v_n}\!\left(\tilde\nabla\tilde v_n\right)
+ \left(-\tilde\mu_n\,\tilde\nabla\tilde v_n\right)\frac{\partial\tilde n}{\partial\tilde v_n}.
\label{eq:gating-chain}
\end{align}
Both terms carry a factor of $\tilde n$. The complete gradient of the continuity loss with respect to $\tilde v_n$ is therefore proportional to $\tilde n$, wherever $\tilde n$ is exponentially small, the gradient is exponentially suppressed irrespective of the magnitude of the residual $r_n$. The same argument applies to $r_p$ via $\tilde p$, and to the Poisson residual $r_P$ via the charge term $\tilde n-\tilde p$. It also applies if $\tilde v_n$ is replaced by $\log\tilde n$ as the network output.

There are two ways to address this shortcoming. First is to circumvent loss minimization, which for example is possible using a Gummel-type iterative solver as used in~\cite{cai_multi-order_2024}, and the second is to include data. We choose to include data as it also helps account for effects not included in the physics equations~\cite{petrosyan_hybrid_2025}, and can allow for generalisation in the future. 

\subsection{Training and evaluation procedure}
\label{sec:training}
The network is trained using Adam optimiser for 20k epochs with an initial learning rate of $10^{-3}$. We drop the learning rate by one order of magnitude every 7k epochs. Additionally we clip gradient norm of each of network at 1.0. We train via drawing a batch of 64 out of the 404 training anchors such that every anchor is visited about 3.2k times. In every epoch each of the 64 anchors contributes 256 collocation points drawn uniformly over trimmed domain. The collocation set is resampled every epoch giving approximately 16k collocation points per step. The 128 contact points of the 64 anchors corresponding to the boundary conditions are fixed. We train the network for six random seeds.

We use the median results over the six seeds for our final analysis to be robust against outliers. For the profiles $(\tilde\phi,\tilde v_n,\tilde v_p)$, which are a function of $x$ we evaluate pointwise median. We checked that the mean and the geometric mean give the same accuracy.

\subsection{Uncertainty estimation framework}
\label{sec:uncertainty-framework}
This section describes our approach to quantify the error in surrogate predictions in absence of ground truth. The recipe in short is as follows. We run the surrogate on devices where we also have the DEVSIM answer, measure how large its errors are, and use those errors to attach an interval to every new prediction. Using our test dataset, we check that the DEVSIM value falls inside the surrogate prediction error bars about 90\% of the time. To do this, we adopt localised conformal-type intervals, validated on held-out test devices with a device bootstrap.

We consider two sources of uncertainty. First, the random seed, which affects the weight initialization, the anchor selection in batch training and the drawing of collocation points. Second, the finite number of devices we sample, in other words, a different device draw would lead to different answers. The two require different treatments. The effect of the seed is reduced by taking the median over six networks. The effect of the finite device sample on the reported statistics is quantified by the device bootstrap, which resamples the test devices.

The seed to seed variation and thus the median over network predictions helps reduce random seed uncertainty, it does not tell us how wrong that prediction is. We establish that error bar using calibration dataset (see sec.~\ref{sec:dataset}), in the spirit of the split conformal prediction framework~\cite{papadopoulos2002inductive,vovk2005algorithmic,lei2018distribution} (see~\cite{angelopoulos2023conformal} for an introduction). 

We empirically observe that the size of the error depends on the depleted device fraction $\kappa_i=\log_{10}(W_{\rm dep}/L)$, where $W_{\rm dep}$ is computed using closed form given in eq.~\eqref{eq:depl_width}. Therefore, we model the error as a function of depletion fraction. 

For a query with a depletion fraction $\kappa_0$, we weight each calibration anchor $i$ by how close its $\kappa_i$ is to $\kappa_0$,
\begin{equation}
w_i(\kappa_0)=\exp\!\left[-\frac{(\kappa_i-\kappa_0)^2}{2h^2}\right],\qquad\qquad h=0.2.
\end{equation}
This allows us to construct an error bar on calibration dataset $\mathcal C$ resulting in relative errors $e_i$ (see eqn.~\ref{eq:QoI-e}), 
\begin{equation}
b(\kappa_0)=\inf\left\{\,b:\ \frac{\sum_{i\in\mathcal C} w_i(\kappa_0)\,\mathbf 1[e_i\le b]}{1+\sum_{i\in\mathcal C} w_i(\kappa_0)}\ \ge\ 1-\alpha\right\},\qquad 1-\alpha=0.90,
\label{eq:errorbar}
\end{equation}
which is the weighted 90\% quantile of the calibration errors, in which the query itself enters the normalisation with weight one and an infinite error, as in weighted conformal prediction. Where the calibration weight near $\kappa_0$ is too small for any $b$ to satisfy the condition, $b=\infty$ and we report no usable bound. This happens outside the calibration support, which is $-2.5\le\log_{10}(W_{\rm dep}/L)\lesssim0.85$ for the current ($W_{\rm dep}/L\lesssim7$), and we refuse to give an interval there. Test queries outside the support are excluded from all coverage and tolerance statements. Weighting the calibration scores by the distance to the query in this way is the idea of kernel-weighted and localised conformal prediction~\cite{lei2014distribution,tibshirani2019conformal,guan2023localized}. Thus for the current the interval
\begin{equation}
\Jtruth\in\left[\frac{\Jpred}{1+b},\ \frac{\Jpred}{1-b}\right]
\label{eq:interval}
\end{equation}
is constructed so that the true current lies within the interval for about 90\% of calibration anchors. It is useful to note that $b$ can be  $\geq 1$, in which case we obtain only one sided error with no upper limit on the interval. We report no usable bound for such a query. 

Now we use the test devices to establish how often eq.~\eqref{eq:interval} holds. This test depends on the set of test devices, therefore we resample the test devices~\footnote{Note that anchors are not independent since they are solutions at different biases and share the same device parameters $n_D, L$.} 2k times with replacement~\cite{field2007bootstrapping,efron1993introduction}. We then compute the coverage
\begin{equation}
\mathrm{coverage}=\frac{1}{|\mathcal T|}\sum_{j\in\mathcal T}\mathbf 1\!\left[e_j\le b(\kappa_j)\right],
\end{equation}
evaluated on all bootstrapped test datasets $\mathcal T$, with $e_j$ the relative error and $b(\kappa_j)$ the error bar defined in eq.~\eqref{eq:errorbar}. This leads to a distribution for coverage, from which we report the 2.5–97.5 percentile interval. The construction is called validated for a quantity if this interval contains $0.90$ or lies above it.

The validity of this procedure rests on the calibration devices being an exchangeable draw of the population on which it is applied, and we state how our design meets this. The exchangeable unit is the device, not the anchor: the voltages of a device lie on a regular grid, so its anchors are neither independent nor a random draw. Calibration and test devices are therefore drawn by the same Latin hypercube design, and the bootstrap resamples devices. The localisation variable $W_{\rm dep}/L$ is a many-to-one projection of $(L,n_D,V)$ and is not a sufficient statistic for the error, so conditioning on it does not give coverage device by device. The coverage is marginal, and it holds to the extent that calibration and test distributions agree given $W_{\rm dep}/L$, which matched designs ensure. The weighted quantile carries no finite-sample guarantee and the device-level exchangeability is only approximate, so we treat coverage as something to be measured on the test devices, not assumed. The networks are trained on data together with a physics penalty that is minimised, not enforced, so the physics residual alone does not bound the error of a derived quantity. The calibration attaches an error scale to it.

At query time, a user may specify a tolerance $\tau$ for the accuracy of the predictions. We accept a query if its bar satisfies $b(\kappa_j) \leq \tau$, in which case the surrogate vouches for the answer to within $\tau$. To test if this holds, we use test devices and measure a hit rate $\sum_{j\ \mathrm{accepted}}\mathbf 1[e_j\le\tau]\big/\#\{\text{accepted}\}$ which we expect to be at least about 0.90, because the errors lie below $b\le\tau$ for 90\% of the calibration anchors. 

\section{Benchmark results across geometries}
\label{sec:benchmarks}
This section shows that the formalism of Sec.~\ref{sec:theory} and its quantities of interest are not tied to one geometry or one doping family. Three single devices as mentioned in Sec.~\ref{sec:bench-data} are trained using the same training recipe and evaluated on four held-out voltages to obtain QoIs as defined in sec.~\ref{sec:qoi}. A detailed study of the one-dimensional symmetric diode follows in Sec.~\ref{sec:results}.

Table~\ref{tab:bench-heldout} summarises the mean relative errors for QoIs for the three devices. The largest errors are the $L_1$ error of the hole density $p$ of the symmetric device (13\%) and the stored charge $Q_n$ (11--13\% for all three devices); all other entries are below 8\%. These are means over the four held-out voltages and include the less accurate voltages discussed below. This shows that the framework itself can produce reliable QoIs for a single device at unseen biases within training range. 

\begin{table}[htbp]
\centering\small
\setlength{\tabcolsep}{6pt}
\renewcommand{\arraystretch}{1.2}
\begin{tabular}{@{}|l|c|c|c|@{}}
\hline
Quantity & 1D symmetric & 1D asymmetric & 2D planar\\
\hline
$\psi$ ($L_1$) & 0.42 & 0.49 & 0.11 \\
$n$ ($L_1$) & 1.1 & 0.47 & 0.50 \\
$p$ ($L_1$) & 13 & 7.5 & 5.0 \\
Peak field $\mathcal{E}_{\rm peak}$ & 0.34 & 0.36 & 4.6 \\
Depletion size $W=|D|$ (2D: area) & 1.8 & 2.2 & 0.69 \\
Stored charge $Q_n$ & 11 & 11 & 13 \\
Stored charge $Q_p$ & 0.41 & 1.4 & 1.3 \\
$\Delta E_F$ ($L_1$) & 0.40 & 0.43 & 0.33 \\
Terminal current $\Jpred$ & 5.1 & 2.1 & 1.3 \\
\hline
\end{tabular}
\caption{Mean (over the four held-out voltages) relative error in percent for QoIs. For the current, the mean is over the held-out voltages at which the anode current density is at least $10^{-6}$\,A/cm$^2$ (0.28, 0.44 and 0.60\,V; for the 2D device the current per unit depth divided by the 0.8\,$\mu$m anode window).}
\label{tab:bench-heldout}
\end{table}

\subsection{One dimensional diodes}
\label{sec:results-asym}
The preceding results are for symmetric junctions ($n_A=n_D$). To test whether the same construction carries over to $n_A\neq n_D$ using the recipe of Sec.~\ref{sec:methods} we train on the asymmetric device ($n_A/n_D=10$) on 12 voltages and evaluate at those and at the four held-out voltages (Sec.~\ref{sec:bench-data}).

\begin{figure}[htbp]
  \centering
  \includegraphics[width=0.85\textwidth]{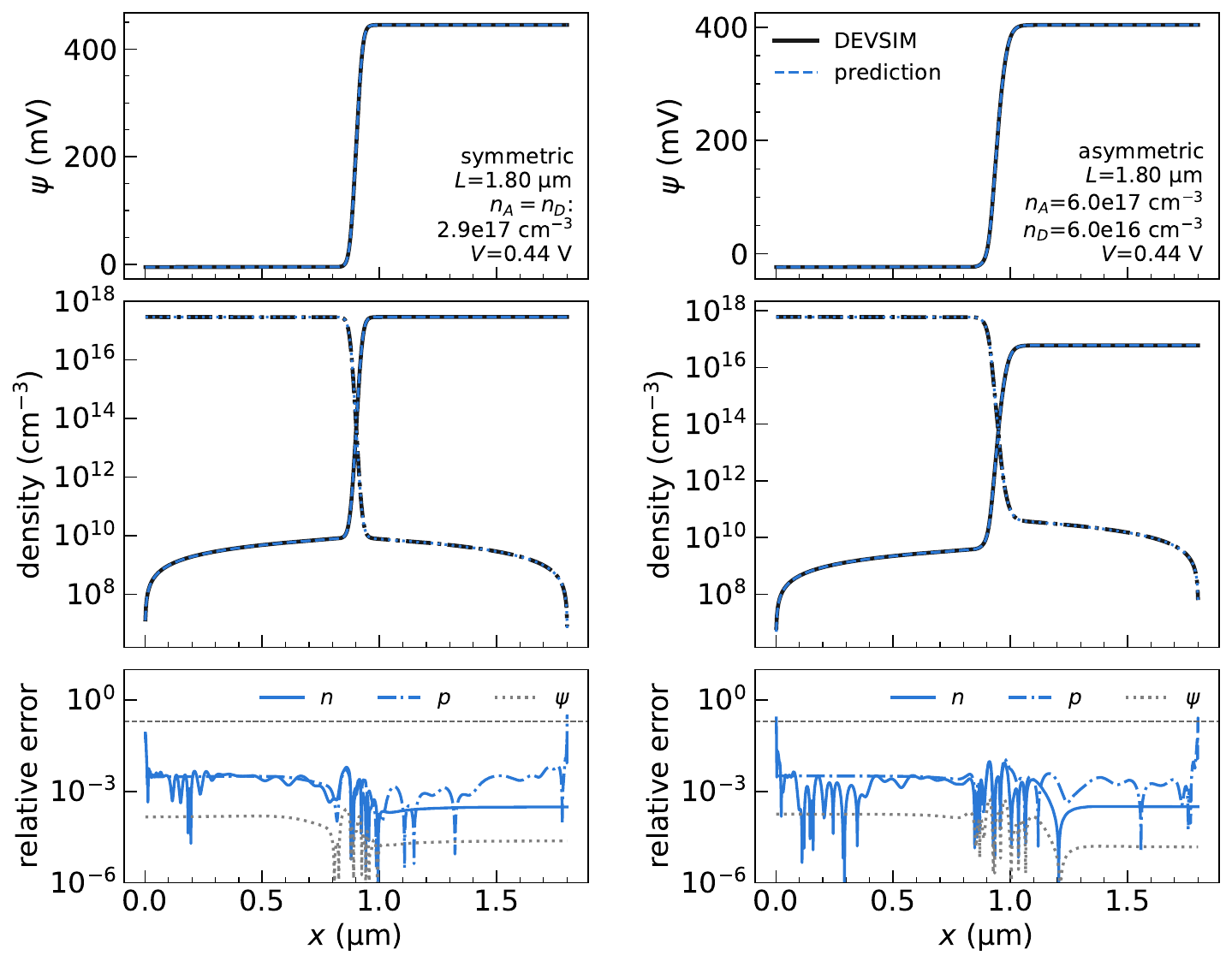}
  \caption{State-space reconstruction of the symmetric (left) and the asymmetric (right) 1D benchmark at the held-out voltage $V=0.44$\,V, on the trimmed domain. Rows: potential (DEVSIM, solid black; network, dashed blue); electron (solid) and hole (dash-dot) densities; and the relative error of $n$, $p$ and $\psi$ (dashed line: 20\%).}
  \label{fig:state-1d}
\end{figure}

Fig.~\ref{fig:state-1d} shows the internal state reconstruction for held-out devices at voltage 0.44\,V. At this voltage the potential is reconstructed across orders of magnitude with residuals below about 0.12\,mV for the symmetric device, and the relative error of the densities remains at the 0.1--1\% level except in the last few nanometres at the contacts, where it reaches 10--30\%. The means in Table~\ref{tab:bench-heldout} are larger because they include the held-out voltages 0.12 and 0.60\,V.

The other two held-out voltages are not reproduced as well (Table~\ref{tab:bench-heldout}). For asymmetric doping, at 0.60\,V the potential RMSE is 5.3\,mV, $Q_n$ is off by 32\% and the current by 6\%; at 0.12\,V, where the current is below the measurable threshold, the depletion width is off by 6\% and $Q_n$ by 13\%. Nevertheless, this shows that the formalism generalises to asymmetric devices as well. 

\subsection{Two-dimensional planar diode}
\label{sec:bench-2d}
\begin{figure}[htbp]
  \centering
  \includegraphics[width=0.95\textwidth]{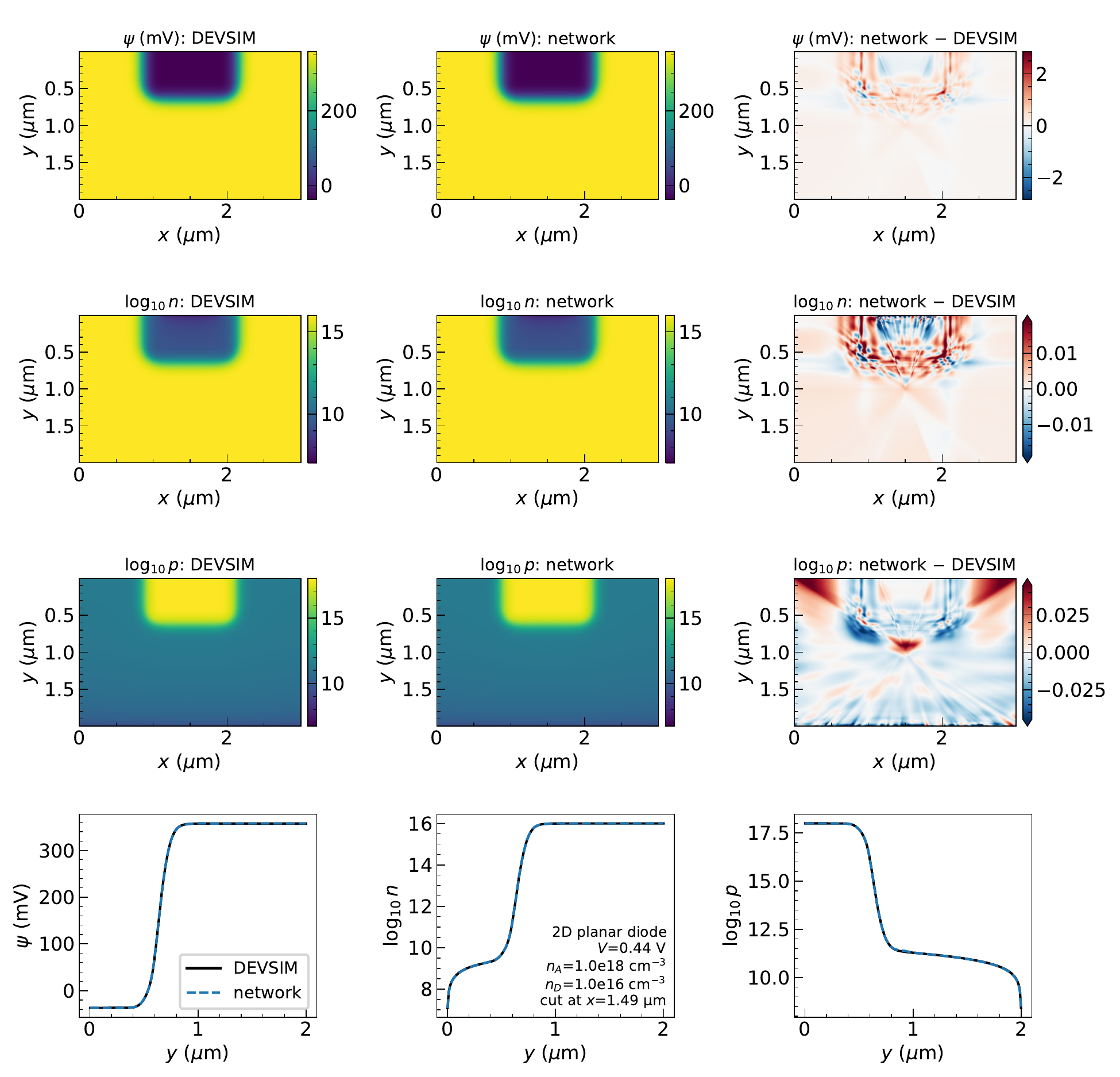}
  \caption{State of the 2D planar benchmark at the held-out voltage $V=0.44$\,V, on the trimmed model box (the anode is at the top, the cathode at the bottom). Rows: potential, $\log_{10}n$ and $\log_{10}p$; columns: DEVSIM, network, and network minus DEVSIM (the colour scale of the two density differences is clipped at the 99th percentile). Bottom: cuts along the vertical line through the middle of the anode window. The potential differs by at most 2.9\,mV, at the corners of the junction. $\log_{10}p$ differs by up to 1.7 in the last row next to the cathode; over the whole box its 99th percentile is 0.04.}
  \label{fig:planar2d-state}
\end{figure}

The 2D device has a curved junction below a narrow anode window and a full-width cathode, so the current enters through a small window and spreads laterally, and the field varies along the rounded junction. Figure~\ref{fig:planar2d-geometry} shows the geometry and the doping. 
Figure~\ref{fig:planar2d-state} shows the state at the held-out voltage 0.44\,V. 

Overall the errors remain small for the internal state reconstruction as well as for the QoI predictions. The largest error is on the stored charge $Q_n$ (13\%, against 1.3\% for $Q_p$), and the peak field is at 4.6\% (Table~\ref{tab:bench-heldout}).

The benchmarks show that our formalism can reproduce the state, the field-based quantities and the derivative-free current across a symmetric and an asymmetric 1D junction and a 2D planar junction. They do not show how far the answers can be trusted or whether they can be generalised over a family of devices. For these questions we concentrate in the rest of the paper on the one-dimensional symmetric diode as a minimal showcase for QoI determination and reliability construction. 

\section{One-dimensional symmetric diode detailed results}
\label{sec:results}
All results in this section and in Sec.~\ref{sec:uncertainty} are for the six-seed ensemble on the 246 test devices, restricted to $V/V_{bi}<0.8$. We discuss one quantity in the main text and delegate supporting tables and figures to Appendix~\ref{app:supplementary}. What the physics term of the loss contributes is tested in Appendix~\ref{app:physics-ablation} by training the same networks without it: in the median the supervised state and the current are as accurate without it, while the peak field and the residuals of the governing equations are clearly worse, and the error distributions have heavier tails. 
\subsection{State and derived quantities}
\label{sec:results-state}
The quantities of interest (defined in Sec.~\ref{sec:qoi}) are the terminal current, depletion width, peak field and stored minority charges, which a designer reads off a device, together with the profiles $\psi$, $n$, $p$, $R$ and $\Delta E_F$ that describe the state they are computed from. Table~\ref{tab:headline-ci} shows the relative error on each QoI for the test, train and held-out devices. The distribution of the errors over devices and voltages is shown in Fig.~\ref{fig:QoI-suite-ensemble} (Appendix~\ref{app:supplementary}). The relative errors on the test devices are larger than on the training devices, by a factor of about 4--6 for $\psi$, $n$ and the current. 
\begin{table}[htbp]
\centering
\renewcommand{\arraystretch}{1.3}
\begin{tabular}{@{}|l|c|c|c|@{}}
\hline
Quantity & Test (246 devices) & Training (27 devices) & Held-out (3 devices) \\
\hline
\multicolumn{4}{@{}l}{Scalar QoIs (relative error)}\\
$\Jpred$ & 0.70 & 0.21 & 1.4 \\
Depletion width $W$ & 1.4 & 0.96 & 1.1 \\
Peak field $\mathcal{E}_{\rm peak}$ & 0.39 & 0.20 & 0.25 \\
Stored charge $Q_n$ & 0.99 & 0.77 & 0.44 \\
Stored charge $Q_p$ & 2.9 & 1.5 & 4.3 \\
\hline
\multicolumn{4}{@{}l}{Profiles (relative $L_1$ error)}\\
$\psi$ & 0.11 & 0.025 & 0.27 \\
$n$ & 1.5 & 0.24 & 4.9 \\
$p$ & 0.61 & 0.27 & 0.68 \\
$R$ & 0.72 & 0.56 & 0.70 \\
$\Delta E_F$ & 0.29 & 0.18 & 0.32 \\
\hline
\end{tabular}
\caption{Median percent error of the ensemble against DEVSIM for the test, training and held-out devices.}
\label{tab:headline-ci}
\end{table}

The errors depend only weakly on the training seed and as shown in table~\ref{tab:seeds}, the ensemble accuracy is comparable to the average single network. Averaging the six networks lowers the median error, and to a large part, the error is shared by the seeds. 

\begin{figure}[htbp]
  \centering
  \includegraphics[width=0.98\textwidth]{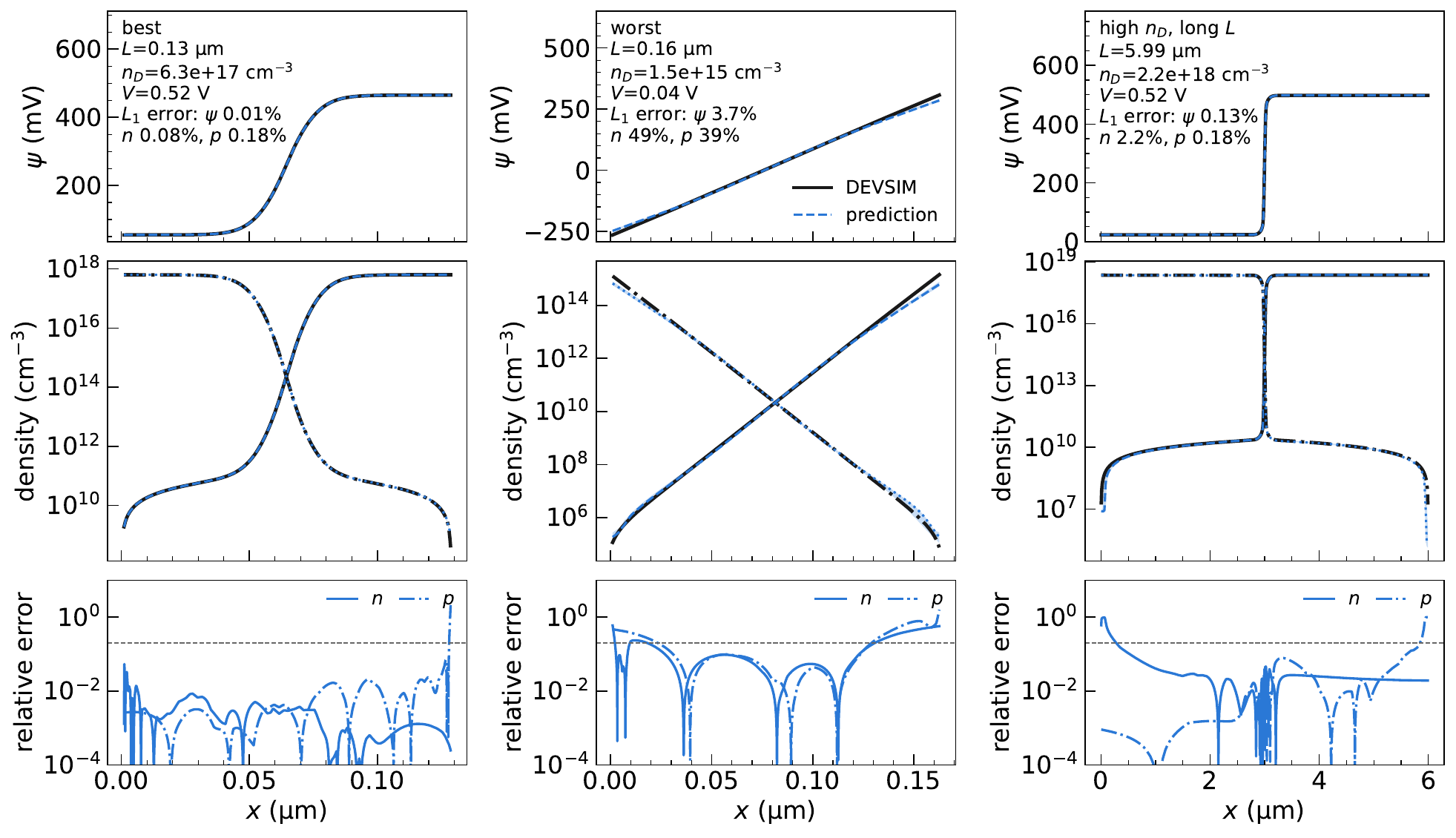}
  \caption{State-space reconstruction on the best and worst mean $L_1$ error for anchors of the unseen devices, and for an anchor of the stiff high-doping, long-device regime.}
  \label{fig:state-space-best-worst}
\end{figure}
Figure~\ref{fig:state-space-best-worst} shows three example state-space reconstructions, the best, worst and one corresponding to high $L, n_D$ device where the solution is particularly stiff. The best reconstruction is indistinguishable from DEVSIM, while for the worst potential is reproduced to within about 22\,mV (5.8--9.0\,mV RMS) and the worst densities have $L_1$ errors of 49\% and 39\% respectively. The  largest errors occur at the contacts. Finally, only the minority tails at the contacts deviate for the stiff anchor. This shows that the network can reconstruct state-space quantities over majority of the regions.

\subsection{Terminal current prediction}
\label{sec:results-readout}
\begin{figure}[htbp]
  \centering
  \includegraphics[width=0.7\textwidth]{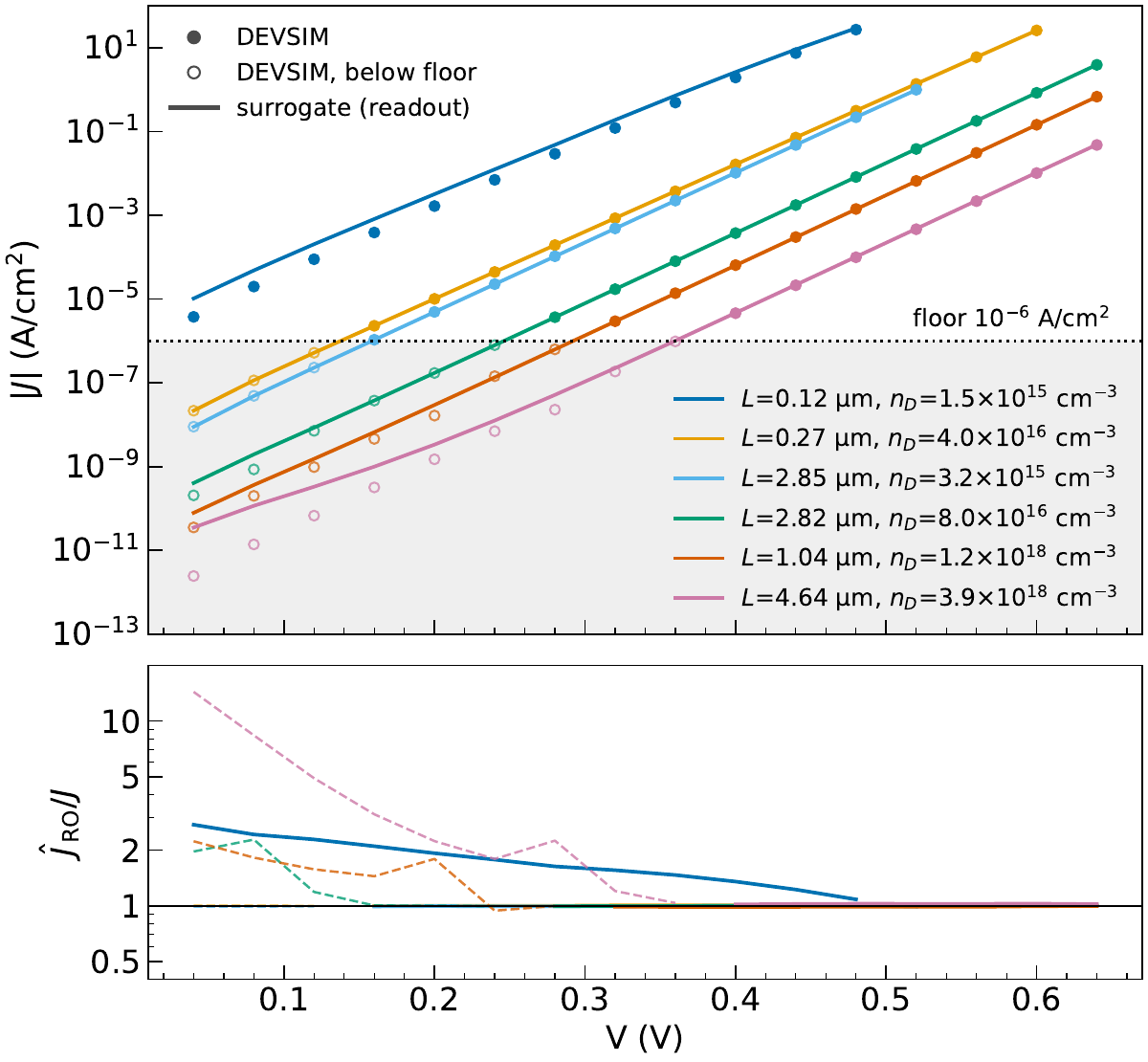}
  \caption{Terminal current against voltage for six test devices spread over current level (top), and the ratio $\Jpred/ \Jtruth$ (bottom); each curve ends at $V/V_{bi}=0.8$ or at 0.64\,V. The shaded region and the open symbols mark points with $|J_{\rm DEVSIM}|<10^{-6}$\,A/cm$^2$; the ratio is dashed below this value. The devices are evenly spaced in mean log-current over the test devices; a device with fewer than five points above the floor or with a wrong-sign readout at any voltage is replaced by the eligible device nearest in current level. Above the floor the current is reproduced over up to seven decades and the largest error occurs at low voltage in the device with $L=0.12\,\mu$m, where the current is small. Below the floor the readout is up to about $14$ times too large (device with $L=4.64\,\mu$m at $0.04$\,V).}
  \label{fig:I-V-response}
\end{figure}

Fig.~\ref{fig:I-V-response} we show the predicted current $\Jpred$ as a function of applied bias for a range of devices. Over most of the regions, the predicted current tracks the DEVSIM truth. The maximal deviation occurs at low bias in long, highly doped devices, where the current is below $10^{-6}$ A/cm$^2$ (dashed). Above this value the largest error is a factor of $2.7$ at $0.04$\,V in a short, lightly doped device.

In anchors with current $>10^{-6}$ A/cm$^2$, the error is particularly correlated with fully depleted devices. In these cases $\Jpred$ defined in eq.~\ref{eq:deriv_free_current} amplifies an error $\delta\tilde\phi$ in the potential as $e^{\delta\tilde\phi}$. Therefore the failure is correlated with the accuracy of $\tilde\phi$ in devices only few Debye lengths long. 
\begin{figure}[htbp]
  \centering
  \includegraphics[width=0.9\textwidth]{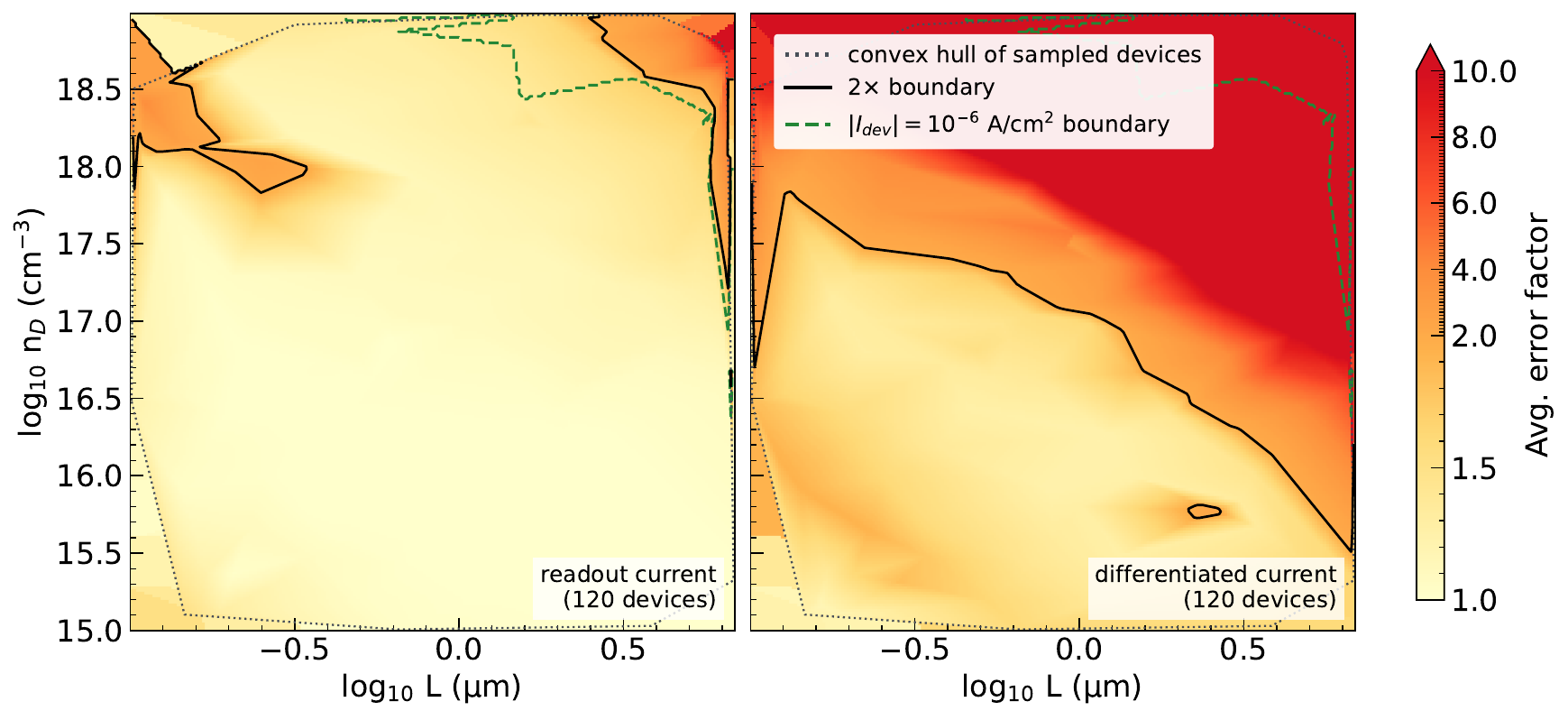}
  \caption{Error factor of the terminal current over the $(L,n_D)$ plane, averaged over voltage, for two different networks. Left: $\Jpred$ of the six-network ensemble of the recipe. Right: $\Jgrad$ from a single modified network, see Appendix~\ref{app:derivative} for details of the modifications. The error factor is $\max(r,1/r)$ for the ratio $r$ of predicted to DEVSIM current.}
  \label{fig:error-map_avg}
\end{figure}

To complete the discussion of current estimation, in fig.~\ref{fig:error-map_avg}, we compare $
\Jpred$ and $\Jgrad$ averaged over all devices and biases. We see that $\Jpred$ is always better than $\Jgrad$ over the entire region, and in particular, $\Jgrad$ can be wrong by more than order of magnitude compared to DEVSIM value specially in large $L$ high $n_D$ region. In appendix~\ref{app:derivative}, we discuss the failure of $\Jgrad$ in detail. 

Given this, we only consider $\Jpred$ for our uncertainty estimates in sec.~\ref{sec:uncertainty}.

\subsection{Extrapolation performance}
\label{sec:results-extrap}
\begin{figure}[htbp]
  \centering
  \includegraphics[width=0.98\textwidth]{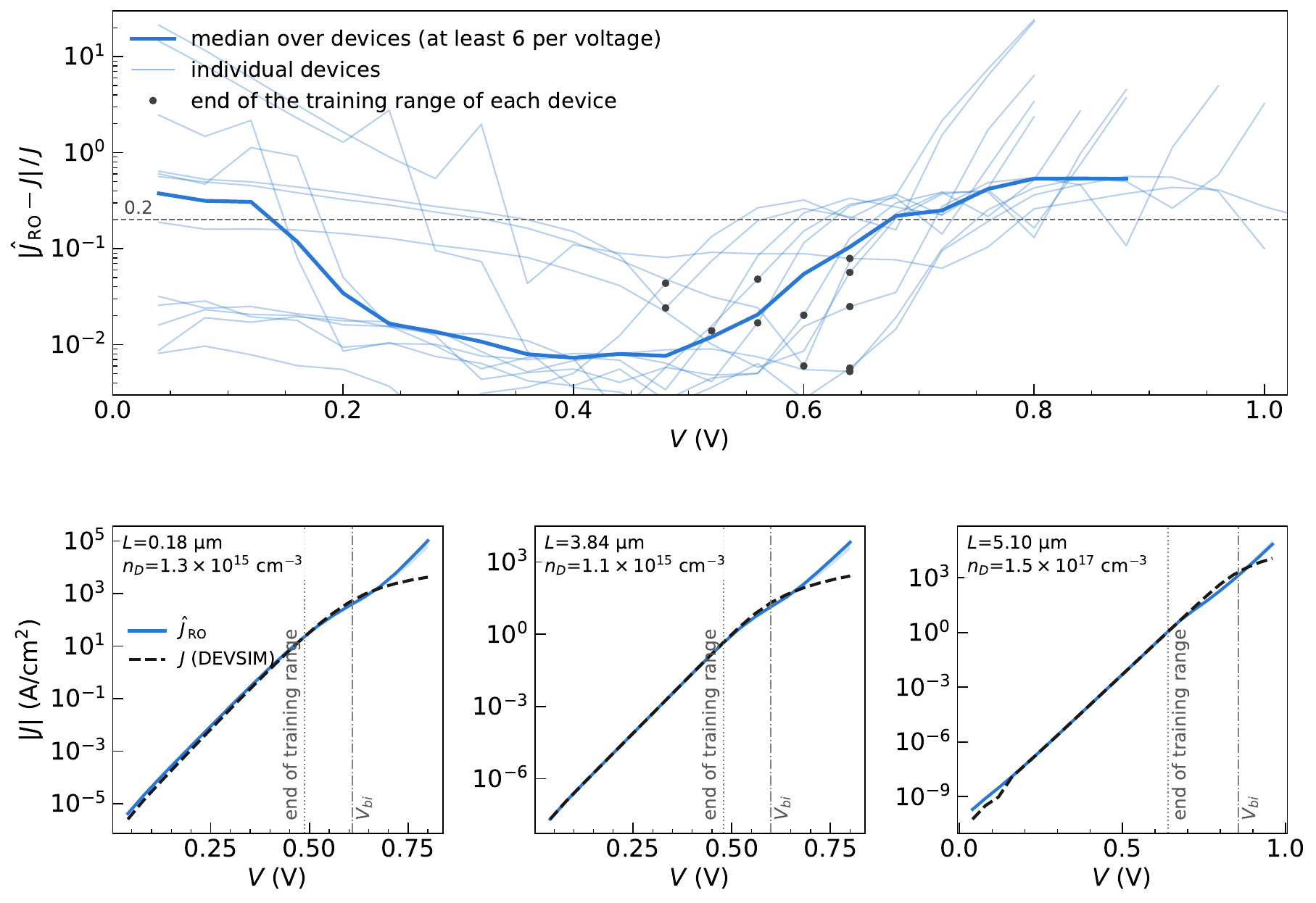}
  \caption{Forward-bias extrapolation beyond the training range, for twelve devices. Top: relative error of the network's terminal current, against $V$ . The large error at low $V$ is the small-current regime and is unrelated to extrapolation.
  Bottom: current against $V$ for the best, median and worst device by median error beyond the end of the training range. The median over devices is drawn where at least six devices contribute.}
  \label{fig:extrap-pilot}
\end{figure}

In Fig.~\ref{fig:extrap-pilot}, we show the six network median error for the extrapolation dataset. Between the end of the training range and $V_{bi}$ the median error of the readout is 25\% against 1.9\% inside the training range. Above $V_{bi}$ the median error is 74\% and the approach fails, where the DEVSIM current becomes resistance-limited and flattens, while the readout keeps growing exponentially. 

The failure above $V_{bi}$ is related to the surrogate conditioning, which is restricted to a range of $(L,n_D,V)$ and is modeled via FiLM~\cite{Perez:2018film}. Beyond their training ranges, the FiLM network extrapolates the response without any restrictions. Therefore, while the physics loss restricts behavior within an anchor, the FiLM response is data-driven. Above $V_{bi}$ the solution changes character, the depletion region collapses and the current becomes limited by the resistance of the quasi-neutral regions. The corresponding profiles are never seen by the network, correspondingly the extrapolation fails. 

\section{Uncertainty quantification for one-dimensional symmetric diode}
\label{sec:uncertainty}
A surrogate delivers a point prediction with no uncertainty attached to it. That prediction is of limited practical value unless the surrogate can also say how wrong it might be, without consulting the ground-truth simulator. We show that an interval conditioned on a single variable covers the true value about ninety percent of the time across all tested regimes. A user can then query the surrogate which not only returns the prediction but also determines if the prediction is within the user-defined tolerance. 

For this, we use conformal predictions as detailed in Sec.~\ref{sec:uncertainty-framework}. Conformal prediction has recently been applied to physics-informed networks and neural PDE solvers~\cite{podina2024conformalized,yu2025conformal,gopakumar2025calibrated,Gopakumar_2026}. Those works calibrate the network outputs. For our application the calibrated object is a scalar quantity derived from the surrogate state.

\begin{figure}[htbp]
  \centering
  \includegraphics[width=0.9\textwidth]{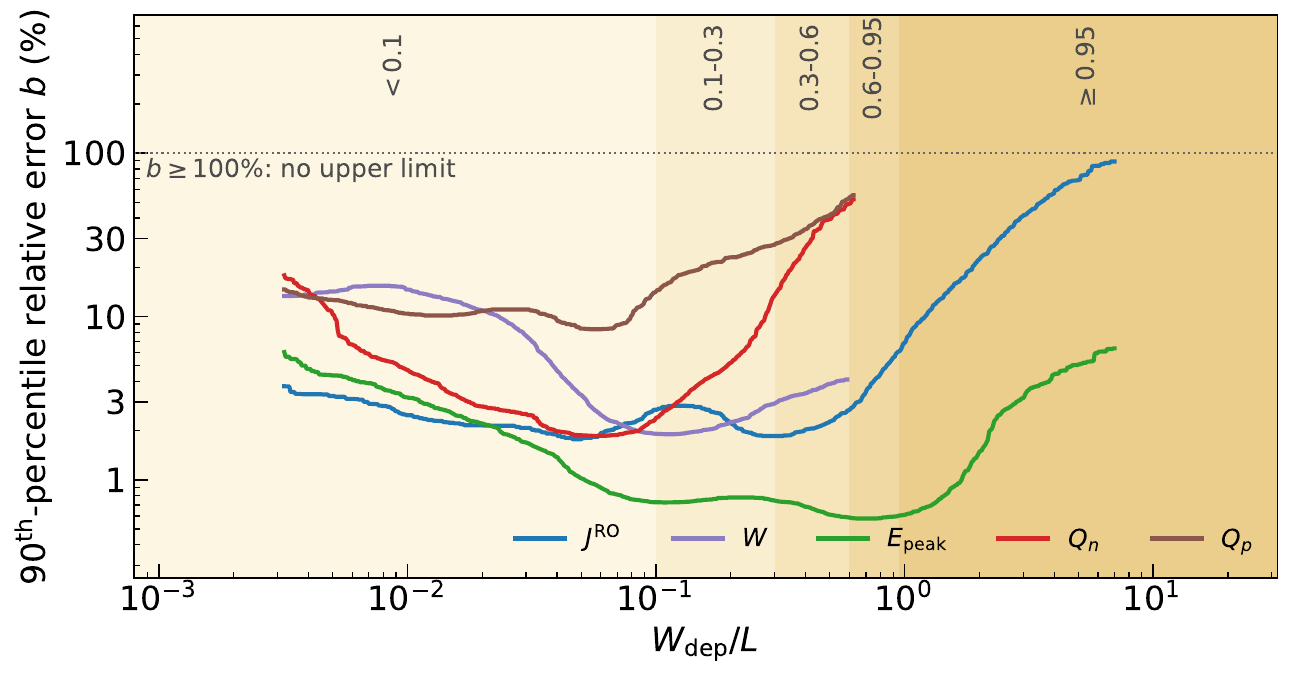}
  \caption{Local 90\% error bar for all five quantities, as a function of depleted fraction. The bar is narrow where the device is quasi-neutral and widens as the depletion region grows. For $W_{\rm{dep}}/L > 0.6$ the stored charges $Q_n$, $Q_p$ and the depletion width are not applicable. The current has a bar up to $W_{\rm{dep}}/L\approx7$; beyond it the interval has no upper limit.}
  \label{fig:calibration_errobar}
\end{figure}

The error of the surrogate is not a fixed fraction of the prediction. A quasi-neutral device, where carrier transport is governed by diffusion across a short junction, is easier for the network than one that is nearly fully depleted, where the current is exponentially sensitive to a potential that must be resolved across a thin space-charge layer. We capture this by conditioning the 90\% error bar (Eq.~\eqref{eq:errorbar}) on the depleted fraction $W_\mathrm{dep}/L$, which is computable in closed form from $(L,n_D,V)$. As Fig.~\ref{fig:calibration_errobar} shows, the bar is narrow across most of the accessible range and widens sharply as the device approaches full depletion. For the current the bar rises to about 0.9 at $W_{\rm dep}/L\approx7$, the edge of the calibration support. Beyond it no bar is available, and the interval has no upper limit.

\begin{figure}[htbp]
  \centering
  \includegraphics[width=0.9\textwidth]{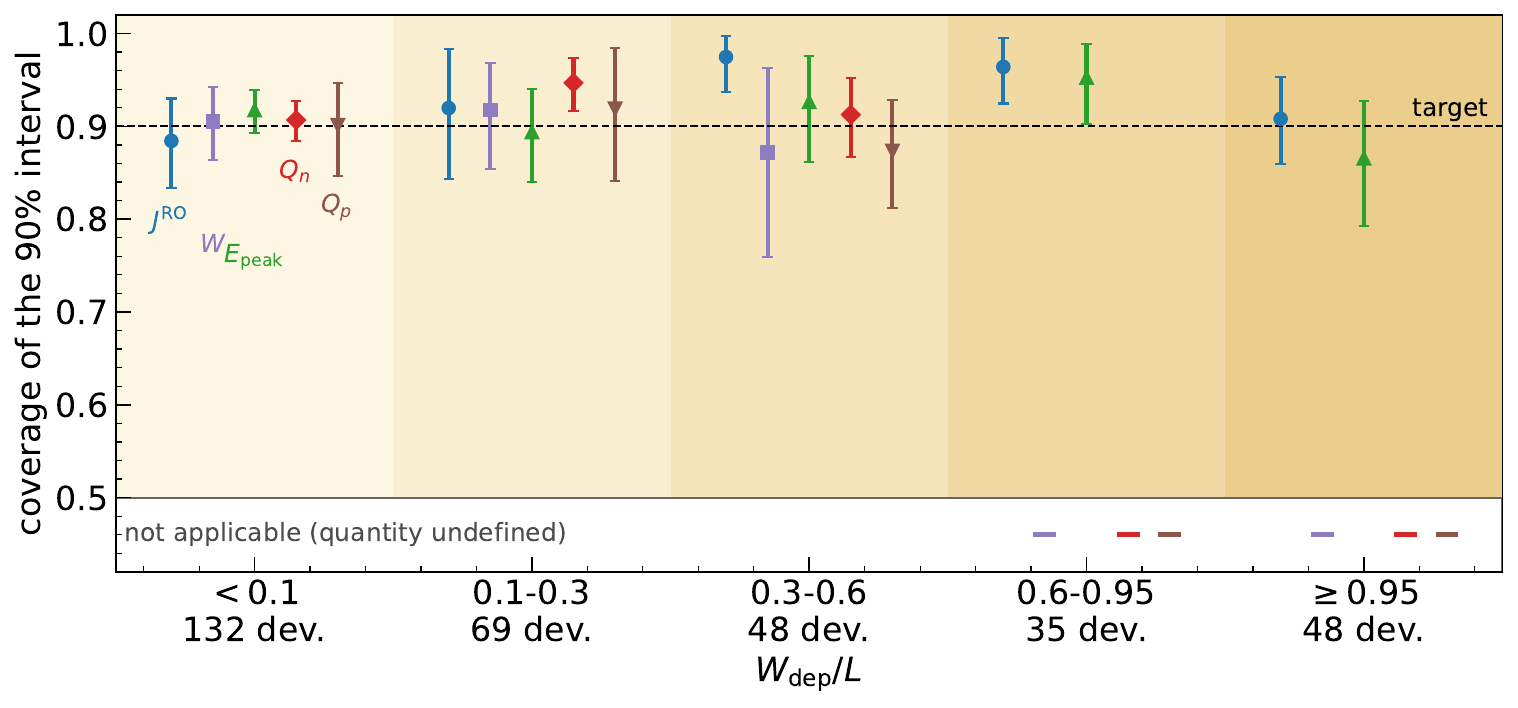}
  \caption{Coverage of the 90\% interval by depleted-fraction regime for the five quantities on the test devices (points: coverage; bars: 95\% device-bootstrap intervals; dashed: target 0.90). A dash marks a regime that is not applicable (fully depleted: the width and the stored charges are undefined).}
  \label{fig:coverage-regime}
\end{figure}

Fig.~\ref{fig:coverage-regime} tests whether this construction holds on unseen devices. The bar was frozen after calibration and then applied to 246 held-out test devices. Coverage is close to the nominal level across the plot. Point estimates below 0.90 are the current at $W_{\rm dep}/L<0.1$ (0.88), the peak field at $0.1$--$0.3$ (0.89) and at $\ge0.95$ (0.87), the depletion width at $0.3$--$0.6$ (0.87) and the hole charge at $0.3$--$0.6$ (0.87). In each case the 95\% interval reaches 0.90. For the current the coverage is 0.96 at $0.6$--$0.95$ and 0.91 at $\ge0.95$. This shows that the bar is small where network is accurate and wide where it is not, thus producing a measure for reliability.

We observe that the range spanned by the six networks covers only 47--73\% of the errors that we obtain here. The ensemble range is not a calibrated predictor, so this is not a comparison with other uncertainty methods. It shows that the seed variation is not a substitute for a calibration, because the networks agree more closely with each other than they agree with DEVSIM. 

While the above results validate our approach, the underlying construction guarantees that nine out of ten queries will be correct. It does not identify which nine. However, a user with a specific tolerance $\tau$ can ask if the surrogate's answer is within the user tolerance.

\begin{table}[htbp]
  \centering
    \renewcommand{\arraystretch}{1.3}
  \begin{tabular}{|l|c|c|c|c|c|c|c|c|}
    \hline
     & & & \multicolumn{2}{c|}{$\tau=0.05$} & \multicolumn{2}{c|}{$\tau=0.10$} & \multicolumn{2}{c|}{$\tau=0.25$} \\
    \cline{4-9}
    QoI & Inform. & Coverage & Acc. & Hit & Acc. & Hit & Acc. & Hit \\
    \hline
    $J$            & 0.69 & $0.91^{+0.03}_{-0.03}$ & 0.80 & $0.96^{+0.02}_{-0.03}$ & 0.85 & $0.98^{+0.01}_{-0.02}$ & 0.92 & $1.00^{+0.00}_{-0.00}$ \\
    $W$            & 0.78 & $0.90^{+0.03}_{-0.03}$ & 0.66 & $0.99^{+0.01}_{-0.01}$ & 0.80 & $1.00^{+0.00}_{-0.00}$ & 1.00 & $1.00^{+0.00}_{-0.00}$ \\
    $E_{\rm peak}$ & 1.00 & $0.91^{+0.02}_{-0.02}$ & 0.99 & $0.99^{+0.00}_{-0.01}$ & 1.00 & $1.00^{+0.00}_{-0.00}$ & 1.00 & $1.00^{+0.00}_{-0.00}$ \\
    $Q_n$          & 0.79 & $0.92^{+0.02}_{-0.02}$ & 0.70 & $0.96^{+0.01}_{-0.02}$ & 0.84 & $0.99^{+0.00}_{-0.01}$ & 0.92 & $0.99^{+0.00}_{-0.00}$ \\
    $Q_p$          & 0.79 & $0.90^{+0.04}_{-0.04}$ & 0.00 & -- & 0.23 & $0.90^{+0.08}_{-0.09}$ & 0.84 & $0.94^{+0.03}_{-0.04}$ \\
    \hline
  \end{tabular}
   \caption{Tolerance statements on the test devices. ``Accepted'': fraction of informative queries whose bar is $\le\tau$; ``hit'': fraction of those whose true relative error is $\le\tau$ (target 0.90; -- : no query accepted). ``Inform.'': fraction of all queries on which the quantity is defined; for $J$ it is the fraction with $|\Jpred|\ge10^{-6}$\,A/cm$^2$, and the other columns refer to these queries.}
  \label{tab:tolerance}
\end{table}

Table~\ref{tab:tolerance} reports, for each quantity, the fraction of queries accepted at each tolerance as well as their hit rate which quantifies the fraction of accepted queries whose true error is indeed within $\tau$. The hit rate target of 0.90 is met for the current and the other QoIs across the considered tolerances, with one exception. For $Q_p$ no query is accepted at $\tau=0.05$, and at $\tau=0.10$ only 23\% are accepted, with a hit rate of 0.90. $Q_p$ is thus usable only at loose tolerances. 

\begin{table}[htbp]
  \centering
  \renewcommand{\arraystretch}{1.3}
\begin{tabular}{|c|c|c|c|c|c|ccccc|}
\hline
$L$ & $n_D$ & $V$ & $W_{\rm dep}/L$ & $J_{\rm pred}$ & $J_{\rm truth}$ & \multicolumn{5}{c|}{$b\le\tau$?\ ($\tau=0.10$)} \\
\cline{7-11}
($\mu$m) & (cm$^{-3}$) & (V) & & (A/cm$^2$) & (A/cm$^2$) & $J$ & $W$ & $E_{\rm peak}$ & $Q_n$ & $Q_p$ \\
\hline
6.9 & $1.81\times10^{18}$ & 0.64 & 0.0032 & $6.96^{+0.27}_{-0.25}\times10^{-2}$ & $7.00\times10^{-2}$ & yes & no & yes & no & no \\
0.91 & $2.65\times10^{18}$ & 0.48 & 0.025 & $7.56^{+0.17}_{-0.16}\times10^{-4}$ & $7.60\times10^{-4}$ & yes & yes & yes & yes & no \\
0.88 & $3.45\times10^{17}$ & 0.64 & 0.05 & $2.97^{+0.05}_{-0.05}$ & $2.97$ & yes & yes & yes & yes & yes \\
0.26 & $1.50\times10^{18}$ & 0.60 & 0.097 & $5.10^{+0.14}_{-0.13}\times10^{-1}$ & $5.12\times10^{-1}$ & yes & yes & yes & yes & no \\
0.15 & $1.41\times10^{18}$ & 0.44 & 0.21 & $2.25^{+0.05}_{-0.05}\times10^{-3}$ & $2.25\times10^{-3}$ & yes & yes & yes & yes & no \\
0.27 & $4.03\times10^{16}$ & 0.52 & 0.49 & $1.38^{+0.03}_{-0.03}$ & $1.38$ & yes & yes & yes & no & no \\
0.13 & $2.94\times10^{16}$ & 0.48 & 1.3 & $2.01^{+0.25}_{-0.20}$ & $1.95$ & no & n.a. & yes & n.a. & n.a. \\
0.12 & $1.46\times10^{15}$ & 0.04 & 8.7 & $1.03\times10^{-5}$ & $3.76\times10^{-6}$ & no & n.a. & no & n.a. & n.a. \\
\hline
\end{tabular}
\caption{Example test queries and their answers: eight measurable test queries at evenly spaced ranks of $W_{\rm dep}/L$. The columns $b\le\tau$ state whether the error bar of the quantity is at most $\tau=10\%$, which needs no DEVSIM value; $J_{\rm truth}$ is shown for comparison. The last query lies outside the calibration support ($W_{\rm dep}/L\gtrsim7$), so no interval is issued.}
  \label{tab:query-answers}
\end{table}

Table~\ref{tab:query-answers} shows eight representative test queries spanning more than three decades of depleted fraction. In the middle of the distribution, where most practical devices operate, the interval is a few percent wide and the ground truth falls comfortably inside it. Toward fully depleted devices the interval opens to a factor of about 1.25 between its bounds ($W_{\rm dep}/L=1.3$), and eventually loses its upper limit, signalling that the surrogate can no longer vouch for the answer. The last query (a 2.7-fold overestimate at $W_{\rm dep}/L = 8.7$) lies outside the calibration support, where no interval is issued. Among the 2511 measurable test anchors inside the calibration support, 91\% lie inside the interval; the other 9 anchors (3 devices) get no interval.

While we show that construction of such intervals is possible, this construction is not without limitations. First and foremost, the coverage we computed is marginal over queries and not a per device quantity, which is the most that can be expected without further assumptions. It also holds only approximately: the localised quantile has no finite-sample guarantee and the device-level exchangeability of calibration and test devices is an approximation, so the coverage is established by the test devices and not proven. Third, intervals are offered only inside the calibration support $-2.5\le\log_{10}(W_{\rm dep}/L)\lesssim0.85$. Below it, in the stiff corner of high doping and long devices ($W_{\rm dep}\sim20$\,nm), the errors are 10--20 times larger than in the central bins and only about 1\% of the devices fall there, too few to calibrate, so we report no interval. Fourth, this calibration and interval construction needs labelled devices not used in training. This requires additional simulations. For a 1D device, it is computationally affordable to run DEVSIM multiple times, however as the complexity of the devices increases, these additional simulations may become an extra challenge. 

It is nevertheless one of the simplest methods to construct reliability test for a surrogate. In the future, we will consider extensions of this framework for 2D/3D devices, where we will analyse the above challenges in detail. 

\section{Conclusions}
\label{sec:conclusion}
We presented a physics-informed surrogate for $p$--$n$ junction diodes in which the network represents the electrostatic potential and the two quasi-Fermi levels. The formalism and the derivative-free current readout apply to one- and two-dimensional devices, as shown for a symmetric and an asymmetric 1D junction and a 2D planar junction, and from one internal state they deliver several quantities of interest for device physicists.

Going beyond these benchmarks, we built a detailed surrogate of a one-dimensional symmetric abrupt $p$--$n$ diode in forward bias, conditioned on the doping, the device length and the applied bias. We used it to predict five quantities of interest, namely, the terminal current, the depletion width, the peak electric field and the stored charges $Q_n$ and $Q_p$.

The quantities of interest are themselves not definition free. The terminal current can be computed by differentiating the network output or by a derivative-free readout based on the Gummel relation, and the readout is the better choice. The differentiated current has to follow a quasi-Fermi slope that varies over many decades across a thin junction layer, which is hard for the network to resolve, whereas the integral of the potential is accurate.

We also constructed localised conformal-type error intervals for each quantity of interest, with the error calibrated as a function of a single variable, the depleted fraction $W_{\rm dep}/L$, on 310 calibration devices drawn like the test devices. On 246 unseen test devices, drawn after everything else was frozen, the marginal coverage is between 0.90 and 0.92 for all five quantities, with a few regimes below nominal whose confidence intervals reach 0.90. The coverage is marginal, not per device, and the intervals are only offered within the calibration support. This turns the surrogate from a predictor into an uncertainty-aware one.  At query time, a user can decide whether the surrogate's answer is accurate to within a tolerance of their choice.

The surrogate is reliable within its training range. Extrapolating in voltage degrades beyond that range, to a median error of 25\% up to the built-in potential against 1.9\% inside it, and it fails above the built-in potential, where the device enters a regime absent from the training data.

The scope of the study is limited. The detailed analysis covers one doping family, a symmetric abrupt junction with a fixed relative smearing of the doping, constant mobility, long lifetimes and forward bias only. Although the benchmarks demonstrate the formalism for non-trivial geometries for a single device, they do not show how well the network generalizes across a family of devices.

Several challenges remain. It is open whether the mesh-free training approach carries over to conditioning on 2D/3D device geometry, where geometry-aware architectures such as graph neural networks may be preferable. Furthermore, more distribution-aware uncertainty models could reduce the need for additional labelled simulations for calibration, at the price of modelling assumptions, and it is of interest which of the two approaches is preferable as device complexity grows.

Despite the limited scope, we believe this is a successful demonstration of a physics-informed framework that returns not only the internal device state but also several quantities of interest, each with a reliability estimate.
\section{Acknowledgements}
We thank the University of Graz for support with the HPC computing. Additionally, we thank Andreas Windisch for useful discussions on the topic.  

During the preparation of this work, we used generative AI for fast experimentation, generate and refine code, and improve the clarity of the manuscript. AI-generated suggestions informed aspects of the methodological development and analysis, which were subsequently critically evaluated, refined, and validated by the author. The author takes full responsibility for the scientific content, computational results, and conclusions presented in this work.
\bibliography{refs}
\appendix
\section{Drift-diffusion theory}
\label{app:dd-theory}
This appendix collects the equations that Sec.~\ref{sec:theory} uses in non-dimensional form. In physical units the problem is
\begin{align}
& \nabla\!\cdot\!\big(\varepsilon\nabla\psi\big) + q\big(p-n+N_{\mathrm{net}}\big) = 0, \label{eq:formalism-poisson}\\
& \nabla\!\cdot\!\mathbf J_n - qR = 0, \qquad\qquad\qquad\quad \nabla\!\cdot\!\mathbf J_p + qR  = 0, \\
& \mathbf J_n  = q\mu_n n \mathbf E + q D_n \nabla n \qquad\qquad
\mathbf J_p  = q\mu_p p \mathbf E - q D_p \nabla p \label{eq:formalism-continuity} \\
& \mathbf E = -\nabla\psi \label{eq:formalism-Efield}
\end{align}

where $\psi$ is the electrostatic potential, $N_{\mathrm{net}}(x)=N_D(x)-N_A(x)$ the net doping (signed step, $-n_A$ for $x<x_j$, $+n_D$ for $x>x_j$), and
\begin{equation}
R = \frac{np-n_i^2}{\tau_p(n+n_i)+\tau_n(p+n_i)}
\label{eq:formalism-srh}
\end{equation}
the SRH rate ($n_i$ the intrinsic concentration, $\tau_n,\tau_p$ carrier
lifetimes). The electron and hole current here is a sum over two terms indicating the drift and diffusion processes and for a successful physics solver they balance exactly in order to yield currents which vary over orders of magnitude as the device parameters and bias ($n_A, n_D, L, V$) change. 

Rather than demanding such balance from the neural network solutions, we adopt a useful well-known reformulation known as the quasi-Fermi level (QFL) parametrization. In QFL parametrization, the state-space vectors $(\psi,n,p)$ are rewritten in terms of electron and hole quasi-Fermi potentials $\phi_n,\phi_p$ and are defined through the Boltzmann relations
\begin{equation}
n = n_i\exp\!\left(\frac{\psi-\phi_n}{U_t}\right), \qquad
p = n_i\exp\!\left(\frac{\phi_p-\psi}{U_t}\right), 
\label{eq:formalism-boltzmann}
\end{equation}
with $U_t = kT/q$. The currents follow as
\begin{equation}
\mathbf J_n = -q\mu_n n \nabla\phi_n, \qquad \mathbf J_p = -q\mu_p p \nabla\phi_p.
\label{eq:formalism-currents}
\end{equation}
It is evident that the current, although algebraically equivalent to the drift-diffusion form, now does not require balancing of terms involving derivatives. 

Although we do not use it in our training, for a planar abrupt junction (the one-dimensional device) a characteristic length scale is the depletion width, which can be approximated as
\begin{equation}
W_{\mathrm{dep}}(V)=\sqrt{\frac{2\varepsilon\,(V_{bi}-V)}{q}\Big(\frac1{n_A}+\frac1{n_D}\Big)},\qquad V_{bi}=U_t\ln\frac{n_A n_D}{n_i^2},
\label{eq:depl_width}
\end{equation}
which reduces to $\sqrt{4\varepsilon(V_{bi}-V)/(qn_D)}$ with $V_{bi}=2U_t\ln(n_D/n_i)$ for $n_A=n_D$. The depletion region extends over $x_p=W_{\rm dep}\,n_D/(n_A+n_D)$ on the $p$-side and $x_n=W_{\rm dep}\,n_A/(n_A+n_D)$ on the $n$-side of the junction, so mostly into the lighter-doped side; for $n_A\gg n_D$ (one-sided junction) $W_{\rm dep}\to x_n=\sqrt{2\varepsilon(V_{bi}-V)/(qn_D)}$.

The non-dimensional quantities of Sec.~\ref{sec:theory} are defined by
\begin{align}
\tilde\phi =& \psi/U_t, \qquad \tilde v_n = \phi_n/U_t  = \tilde\phi-\ln(n/n_i) \qquad \tilde v_p=\phi_p/U_t = \tilde\phi+\ln(p/n_i) \\
\mu_{\max} &= \max(\mu_n,\mu_p), \qquad
\lambda_D = \sqrt{\varepsilon U_t/(qn_{\max})}, \qquad \lambda = \lambda_D/\ell,\\
\tilde\mu_n &= \mu_n/\mu_{\max}, \quad \tilde\mu_p=\mu_p/\mu_{\max}, \\
\tilde n_i & = n_i/n_{\max}, \quad \tilde\tau_n = \tau_n\mu_{\max} U_t/\ell^2, \quad
\tilde\tau_p = \tau_p\mu_{\max} U_t/\ell^2,\\
J_{\mathrm{ref}} &= \frac{q\mu_{\max} U_t n_{\max}}{\ell}, \qquad
R_{\mathrm{ref}} = \frac{\mu_{\max} U_t n_{\max}}{\ell^2}, \qquad n_{\max}=\max(n_A,n_D).
\label{eq:formalism-scaling}
\end{align}

\section{Error distributions and seed variation}
\label{app:supplementary}
These tables and figures support results stated in Sec.~\ref{sec:results} and Sec.~\ref{sec:uncertainty}. Each repeats or extends an exhibit that carries the claim in the main text.
\begin{figure}[htbp]
  \centering
  \includegraphics[width=\textwidth]{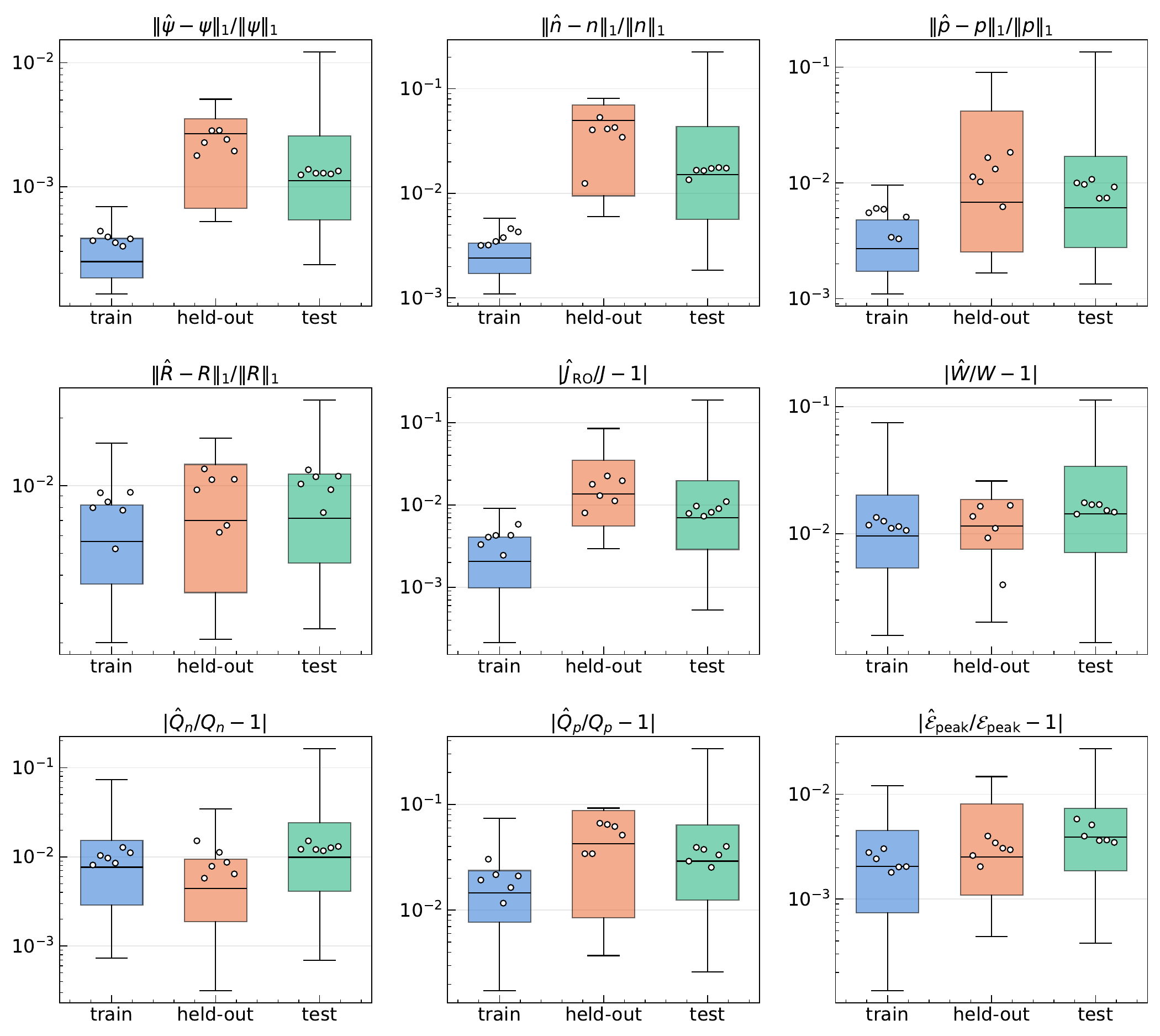}
  \caption{Distribution of the ensemble error (Eq.~\eqref{eq:QoI-e}; relative $L_1$ error for the profiles in the top row, relative error for the scalar QoIs in the bottom row) over the (device, voltage) points of the training, held-out and test devices, on a logarithmic axis. Boxes: quartiles with median; whiskers: 5th and 95th percentiles. Dots: the median error of each of the six single networks on that set, so the spread of the dots around the box is the variation between networks. The held-out set has three devices and is indicative only. Table~\ref{tab:headline-ci} gives the test medians.}
  \label{fig:QoI-suite-ensemble}
\end{figure}

\begin{table}[htbp]
\centering
\renewcommand{\arraystretch}{1.2}
\begin{tabular}{@{}|l|c|c|c|@{}}
\hline
 & Ensemble error & Single network & Single network \\
Quantity & & mean $\pm$ std & min--max \\
\hline
\multicolumn{4}{@{}|l|}{Scalar QoIs (relative error)}\\
Readout current $\Jpred$ & $0.70^{+0.14}_{-0.10}$ & 0.88 $\pm$ 0.13 & 0.73--1.1 \\
Depletion width $W$ & $1.4^{+0.24}_{-0.19}$ & 1.6 $\pm$ 0.14 & 1.4--1.8 \\
Peak field $\mathcal{E}_{\rm peak}$ & $0.39^{+0.043}_{-0.037}$ & 0.43 $\pm$ 0.095 & 0.35--0.58 \\
Stored charge $Q_n$ & $0.99^{+0.15}_{-0.14}$ & 1.3 $\pm$ 0.12 & 1.2--1.5 \\
Stored charge $Q_p$ & $2.9^{+0.56}_{-0.45}$ & 3.4 $\pm$ 0.59 & 2.5--4.0 \\
\hline
\multicolumn{4}{@{}l}{Profiles (relative $L_1$ error)}\\
$\psi$ & $0.11^{+0.019}_{-0.012}$ & 0.13 $\pm$ 0.0050 & 0.12--0.14 \\
$n$ & $1.5^{+0.37}_{-0.26}$ & 1.6 $\pm$ 0.15 & 1.3--1.8 \\
$p$ & $0.61^{+0.091}_{-0.072}$ & 0.91 $\pm$ 0.14 & 0.74--1.1 \\
$R$ & $0.72^{+0.054}_{-0.056}$ & 1.0 $\pm$ 0.15 & 0.76--1.2 \\
$\Delta E_F$ & $0.29^{+0.031}_{-0.028}$ & 0.37 $\pm$ 0.049 & 0.31--0.44 \\
\hline
\end{tabular}
\caption{Dependence on the random seed. The six networks differ only in their seed and have a common training dataset.}
\label{tab:seeds}
\end{table}

For every QoI the error of the ensemble is below the mean error of a single network. The readout current is the quantity least affected by the seed (single-network errors 0.73--1.1\%). The ensemble error lies below the whole range of the single networks for the readout current, $Q_n$, $\psi$, $p$, $R$ and $\Delta E_F$. For the other QoIs the best single network is as accurate as the ensemble or slightly better (for example 0.35\% against 0.39\% for the peak field).
\section{$\Jgrad$ analysis}
\label{app:derivative}
In this appendix we investigate the origin of the failure of $\Jgrad$ to obtain current. In order to assist the network, we add two more features to the training pipeline. First the built-in-potential column $\eta=(V_{bi}-V)/U_t$ in the conditioning vector. This is added such that the network is aware of how far it is from the built-in potential, where the current becomes resistance limited. Secondly, the size of the continuity residual differs strongly between anchors. In the scaled units the peak SRH recombination rate, which sets this size, is proportional to
\begin{equation}
	R_{\mathrm{peak,ref}}=\frac{n_{D,\mathrm{ref}}/n_{\max}}{2\sqrt{\tilde\tau_n\tilde\tau_p}}\,e^{V/2U_t},
\end{equation}
with $n_{D,\mathrm{ref}}=10^{15}\,\mathrm{cm^{-3}}$. It falls as $n_{\max}$ increases and grows with the bias. As we train over a pool of anchors, the continuity loss therefore varies strongly from anchor to anchor, which is a mechanism similar to the density gating. To compensate, we divide the continuity residuals by $R_{\mathrm{peak,ref}}$. This brings the losses of all anchors to a similar magnitude and allows for a better optimisation. The physical peak rate has $n_i$ in place of $n_{D,\mathrm{ref}}$, so $R_{\mathrm{peak,ref}}$ differs from it by the constant factor $n_{D,\mathrm{ref}}/n_i$, a numerical choice. 

With these two additions, we train a single network and evaluate it on the measurable anchors of test devices of an earlier Latin hypercube draw. On these anchors the differentiated current is within a factor of two of DEVSIM for 72\% of the anchors, against 98\% for the readout. These two features were introduced to give $\Jgrad$ its best chance, so this comparison favours $\Jgrad$ and yet falls short of delivering a high accuracy answer. Without the additional features, only 13\% of the anchors lead to a current within a factor of two. We deliberately choose a loose acceptance criteria to demonstrate the $\Jgrad$ weakness.

\begin{figure}[htbp]
  \centering
  \includegraphics[width=0.98\textwidth]{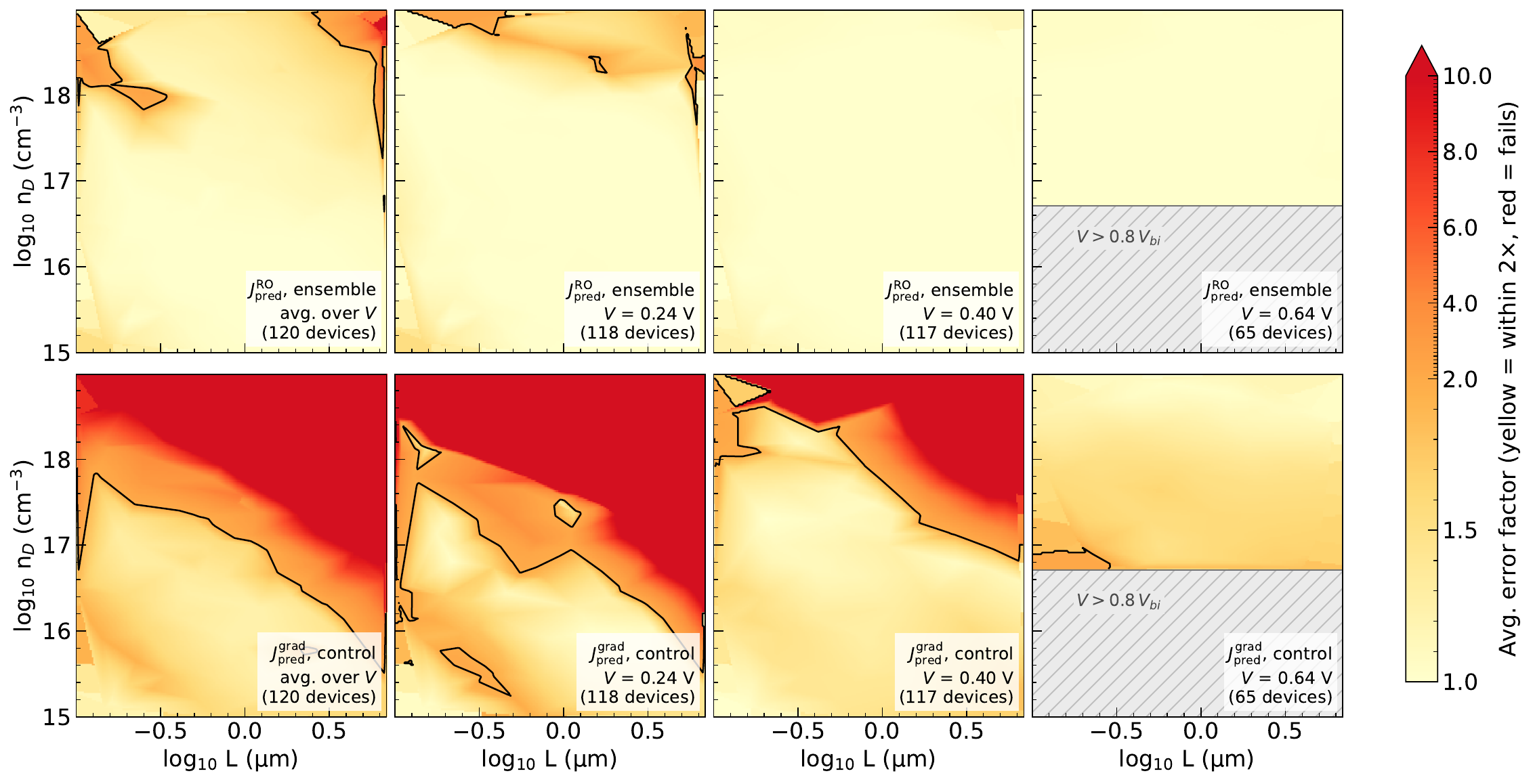}
  \caption{Error factor of the readout current $\Jpred$ of the ensemble (top row) and of the differentiated current $\Jgrad$ of the network with the $\eta$ column and the $R_{\rm peak,ref}$ divisor (bottom row) over the $(L,n_D)$ plane, for the same 120 unseen devices, including those with $|J_{\rm truth}|<10^{-6}$\,A/cm$^2$. The hatched region at $0.64$\,V marks the devices with $V\ge0.8\,V_{bi}$, which are excluded.}
  \label{fig:error-map-derivative}
\end{figure}

Figure~\ref{fig:error-map-derivative} compares the best readout with the best differentiated current. The top row is the readout $\Jpred$ of the ensemble network. The bottom row is the differentiated current $\Jgrad$ of the network with the $\eta$ column and the $R_{\mathrm{peak,ref}}$ divisor. The two rows are different networks by design and use the same 120 unseen devices, and the figure compares the broad pattern of failure. Unlike the rest of this appendix, the maps include anchors with $|J_{\rm truth}|<10^{-6}$\,A/cm$^2$, where the DEVSIM current is close to its numerical floor, so part of the red region at the lowest voltages may reflect that floor. The readout is within a factor of two over nearly the whole plane at every voltage shown, with small regions of larger error at the highest doping. This corner shrinks as the voltage grows and is within a factor of two at $0.64$\,V. Anchors with $V/V_{bi}\ge0.8$ ($V_{bi}=2U_t\ln(n_D/n_i)$) are excluded, so at $0.64$\,V only devices with $n_D\gtrsim5\times10^{16}$\,cm$^{-3}$ remain (65 devices) and the band of intermediate doping that fails above $0.8\,V_{bi}$ is not shown.

We find the failure to be associated with the device length in units of the Debye length, $L/\lambda_D$, and offer the following interpretation. A device that is many Debye lengths long is mostly quasi-neutral, and the carrier densities change by many decades across a thin layer around the junction. Neglecting the small recombination current, the electron current $J_n=-q\mu_n n\,\mathrm dv_n/\mathrm dx$ has to be the same everywhere in the device, so the slope $\mathrm dv_n/\mathrm dx$ has to follow $1/n$, that is change by many decades within a few Debye lengths. An error of the slope is further multiplied by the density, which amplifies the error. 

\begin{figure}[htbp]
  \centering
  \includegraphics[width=0.98\textwidth]{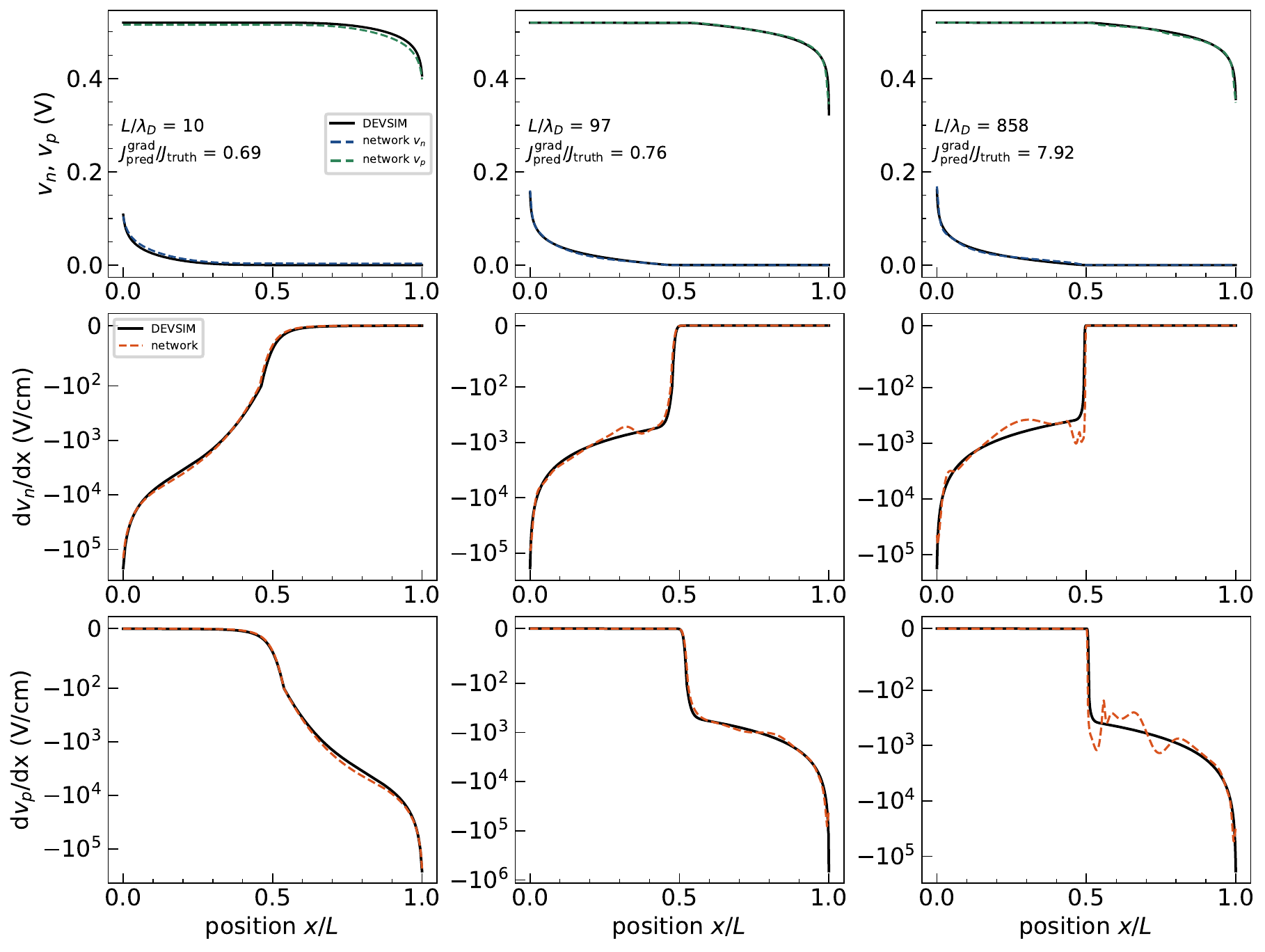}
  \caption{Quasi-Fermi levels $v_n$ and $v_p$ (top row) and their slopes $\mathrm dv_n/\mathrm dx$ (middle) and $\mathrm dv_p/\mathrm dx$ (bottom) along three unseen devices at $V=0.52$\,V, of the network with the $\eta$ column and the SRH-peak divisor, for $L/\lambda_D$ of 10, 97 and 858. Black: DEVSIM; dashed: network. The slope axes are symmetric logarithmic (linear below 100\,V/cm). Each column is labelled with $L/\lambda_D$ and $J^{\rm grad}_{\rm pred}/J_{\rm truth}$.}
  \label{fig:quasi-fermi-slopes}
\end{figure}

Figure~\ref{fig:quasi-fermi-slopes} shows the quasi-Fermi levels and their slopes along the three devices, for $L/\lambda_D=10$, 97 and 858. While the agreement between DEVSIM and network predictions is excellent for $v_n$ and $v_p$ their slopes show a different picture. The slope of the network follows that of DEVSIM at $L/\lambda_D=10$ and 97 with small deviations, while at 858 it oscillates around it in the majority-carrier bulk on both sides of the junction, with amplitudes comparable to the slope itself. We have not established the origin of these oscillations, but for these three devices the figure illustrates that the $v_n, v_p$ slopes degrade as $L/\lambda_D$ increases. 

\section{Ablation of the physics term}
\label{app:physics-ablation}
This appendix tests the contribution of physics in the final network prediction. To this end, we trained additional networks using the 1D symmetric train dataset with data and boundary term loss only. We train six such networks, with the same six seeds as the physics-informed ones, and score them on the test devices exactly as in the main text. Numbers are medians over the test anchors of this ensemble.

\begin{figure}[htbp]
  \centering
  \includegraphics[width=\textwidth]{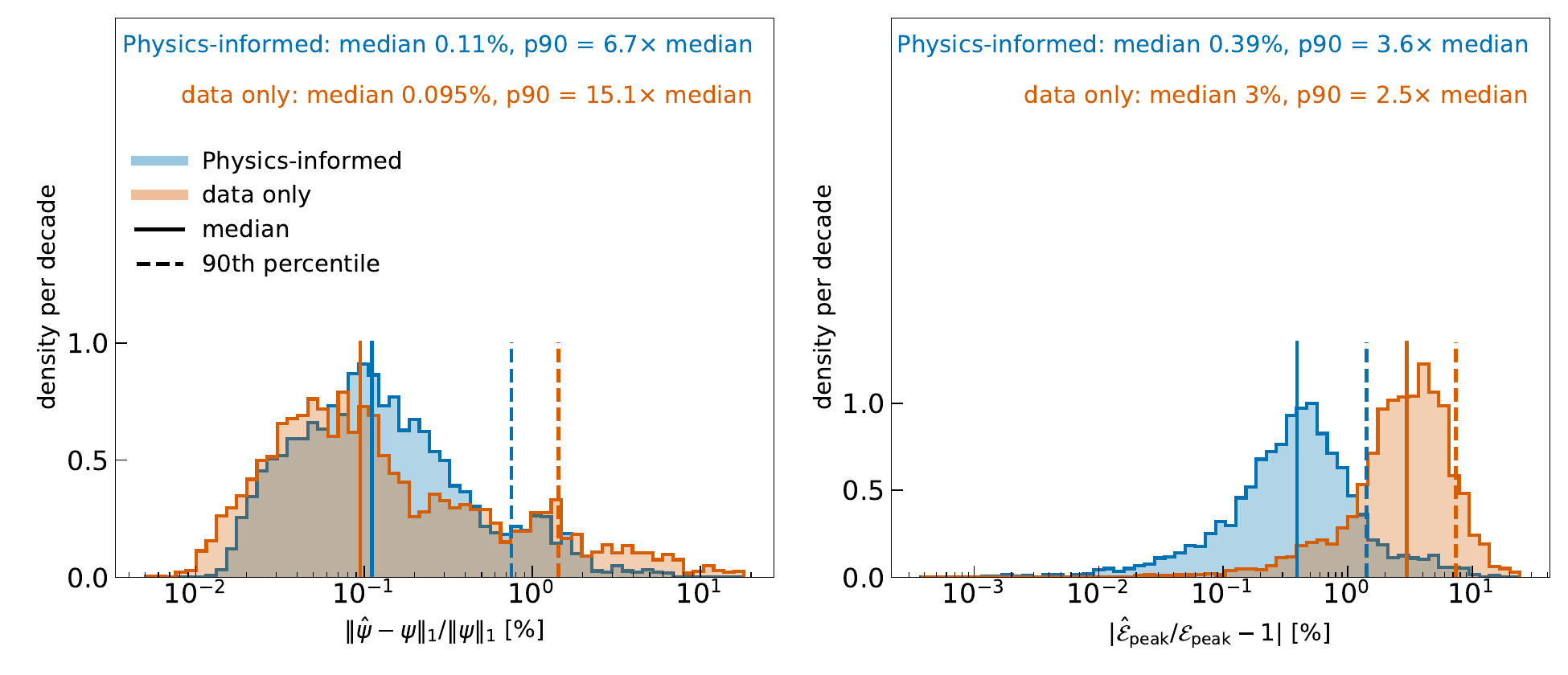}
  \caption{Distribution of the error of each (device, voltage) point of the test devices on a logarithmic axis, for the physics-informed and the data-only six-network ensembles. Left: relative $L_1$ error of $\psi$; right: relative error of the peak field. Solid lines: median, dashed lines: $90$th percentile of each set. The text above each panel gives the median and the ratio of the $90$th percentile to the median. Without the physics term the typical error of $\psi$ is the same and the tail is heavier; the distribution of the error of the peak field is shifted to larger values.}
  \label{fig:physics-ablation}
\end{figure}

We observe that the median of the physics-informed and data network predictions are comparable on the supervised fields and on the current. However, Fig.~\ref{fig:physics-ablation} (left panel) shows that the error distributions of data only network have heavier tails. Relative to its own median, the $90$th percentile of the error on $\psi$, $n$ and $\Jpred$ is about twice as large in the data-only networks as in the physics-informed ones.

The physics-informed advantage is particularly visible in the peak E field as seen in the right panel of Fig.~\ref{fig:physics-ablation}. The error of the peak field $\mathcal{E}_{\rm peak}$ is about $8$ times larger in the data-only ensemble than in the physics-informed one, and the whole distribution is shifted. The relative $L_1$ error of the field of a single network is larger by a factor of $2.3$. The gap thus grows with the order of the derivative of $\psi$ on which a quantity depends: none for the values of $\psi$, $n$ and $p$, about two for the field, and about eight for its peak. The quantities that depend on the field only through the depletion edges, $W$ and $Q_n$, differ little.

We evaluate the residuals of the Poisson and of the two continuity equations at fixed collocation points. Compared with a physics-informed network, the median residual of the data-only networks is larger by a factor of about $60$ for the Poisson equation, about $3000$ for the electron and about $1000$ for the hole continuity equation, and is larger for more than 90\% of the anchors.

We expect a larger difference to arise if the physics model contains process activated by the field, such as impact ionisation. This may favour physics-informed training and will be tested in the future.

\end{document}